\documentclass[lettersize,journal]{IEEEtran}
\usepackage{amsmath,amsfonts}

\usepackage[ruled,linesnumbered]{algorithm2e}
\usepackage{array}
\usepackage[caption=false,font=normalsize,labelfont=sf,textfont=sf]{subfig}
\usepackage{textcomp}
\usepackage{stfloats}
\usepackage{url}
\usepackage{verbatim}
\usepackage{graphicx}
\usepackage{changepage}
\usepackage{cite}
\usepackage{multirow}
\usepackage{amsthm}
\usepackage{amsfonts,amssymb}
\usepackage{booktabs}
\usepackage[table]{xcolor}
\usepackage{color}
\definecolor{r}{rgb}{1,0,0}

\begin{document}

\title{Efficient Generation of Targeted and Transferable Adversarial Examples for Vision-Language Models Via Diffusion Models}

\author{Qi Guo, Shanmin Pang$^{\dagger}$,~\IEEEmembership{Member,~IEEE}, Xiaojun Jia$^{\dagger}$, Yang Liu,~\IEEEmembership{Senior Member,~IEEE}, Qing Guo,~\IEEEmembership{Senior Member,~IEEE}

}

\markboth{IEEE Transactions on Information Forensics and Security}%
{Shell \MakeLowercase{\textit{et al.}}: Efficient Generation of Targeted and Transferable Adversarial Examples for Vision-Language Models Via Diffusion Models}


\maketitle

\begin{abstract}
Adversarial attacks, particularly \textbf{targeted} transfer-based attacks, can be used to assess the adversarial robustness of large visual-language models (VLMs), allowing for a more thorough examination of potential security flaws before deployment. However, previous transfer-based adversarial attacks incur high costs due to high iteration counts and complex method structure. 
Furthermore, due to the unnaturalness of adversarial semantics, the generated adversarial examples have low transferability.
These issues limit the utility of existing methods for assessing robustness. To address these issues, we propose AdvDiffVLM, which uses diffusion models to generate natural, unrestricted and targeted adversarial examples via score matching. Specifically, AdvDiffVLM uses Adaptive Ensemble Gradient Estimation (AEGE) to modify the score during the diffusion model's reverse generation process, ensuring that the produced adversarial examples have natural adversarial targeted semantics, which improves their transferability. Simultaneously, to improve the quality of adversarial examples, we use the GradCAM-guided Mask Generation (GCMG) to disperse adversarial semantics throughout the image rather than concentrating them in a single area. Finally, AdvDiffVLM embeds more target semantics into adversarial examples after multiple iterations.
Experimental results show that our method generates adversarial examples 5x to 10x faster than {state-of-the-art (SOTA)} transfer-based adversarial attacks while maintaining higher quality adversarial examples. Furthermore, compared to previous transfer-based adversarial attacks, the adversarial examples generated by our method have better transferability. Notably, AdvDiffVLM can successfully attack a variety of commercial VLMs in a black-box environment, including GPT-4V. {The code is available at https://github.com/gq-max/AdvDiffVLM}

\end{abstract}


\section{Introduction}\label{sec:1}
\IEEEPARstart{L}{arge} VLMs have shown great success in tasks like image-to-text~\cite{blip,blip2,llava} and text-to-image generation~\cite{ldm, unidiffuser}. Particularly in image-to-text generation, users can use images to generate executable commands for robot control~\cite{robot}, which has potential applications in autonomous driving systems~\cite{vlm2scene, drive_survey}, visual assistance systems~\cite{visual_assistance}, and content moderation systems~\cite{content_moderation}. However, VLMs are highly susceptible to adversarial attacks~\cite{han2023ot,gao2024boosting}, which can result in life and property safety issues~\cite{agent_smith,attack_drive}. As a result, it is critical to evaluate the adversarial robustness~\cite{west2023towards,attackvlm,jia2024revisiting,jia2024improving} of these VLMs before deployment.

The early research on assessing the adversarial robustness of VLMs concentrated on white-box and untargeted scenarios~\cite{image-caption-attack,exact,gray-box}. Black-box and targeted attacks can cause models to generate targeted responses without knowing the models' internal information, resulting in greater harm~\cite{baia2023black,zhu2023efficient}. {Furthermore}, targeted attacks on black-box models present more challenges than untargeted attacks~\cite{williams2023black,zhu2023boosting}. As a result, when assessing the adversarial robustness of VLMs, it is critical to consider more threatening and challenging black-box and targeted attacks~\cite{attackvlm}. AttackVLM~\cite{attackvlm} is the first work to explore the adversarial robustness of VLMs in both black-box and targeted scenarios using query attacks with transfer-based priors. However, due to the large number of queries required and the complex method structure, this method is inefficient, {which reduces its validity and suitability for a comprehensive assessment of the limitations of VLMs.} Another attack method that can be used in black-box and targeted scenarios is the transfer-based attack~\cite{cwa,svre,sia,fap}. However, this type of attack method is slow to generate adversarial examples due to its complex structure and numerous iterations. {Furthermore}, because it adds unnatural adversarial semantics, the transferability of adversarial examples is poor. Unrestricted adversarial examples~\cite{uae1,uae2,advdiffuser,colorfool} can incorporate more natural adversarial targeted semantics into the image, thereby improving the image quality and transferability of adversarial examples. For example, AdvDiffuser~\cite{advdiffuser} incorporates PGD~\cite{pgd} into the reverse process of the diffusion model to generate targeted adversarial examples with better transferability against classification models. However, applying PGD to the latent image in the reverse process is not suitable for the more difficult task of attacking VLMs. At the same time, performing PGD on each step of the reverse process incurs high costs. 

\begin{figure*}[t]
\centering\includegraphics[width=\linewidth]{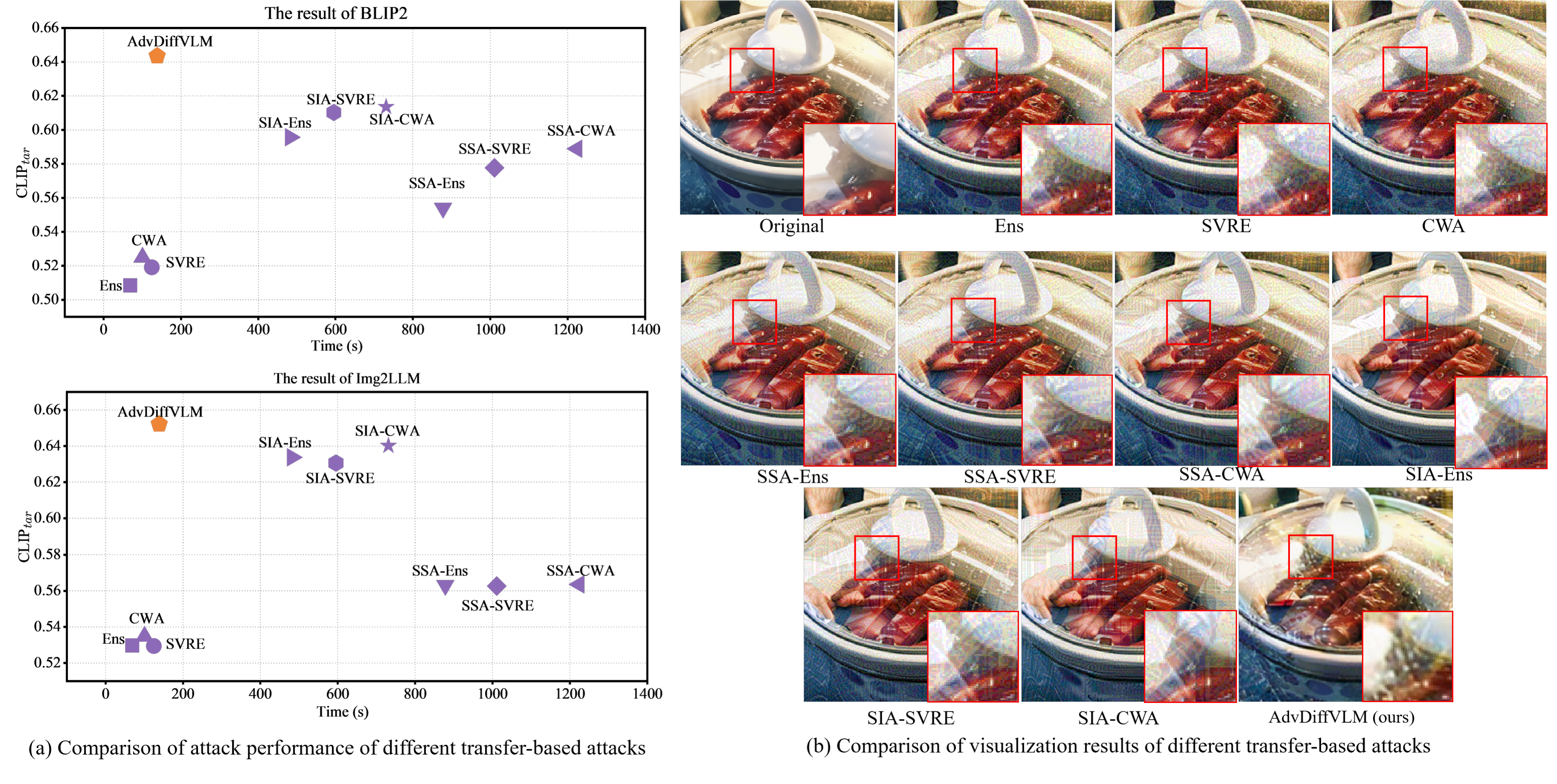}
\vskip -0.05in
\caption{Comparison of different transfer-based attacks and our method on VLMs. (a) Comparison of attack performance. We select BLIP2~\cite{blip2} and Img2LLM~\cite{image2llm} as the representation models of VLMs. We select existing transfer-based attacks  in conjunction with AttackVLM~\cite{attackvlm} as comparison methods, including Ens~\cite{mifgsm}, SVRE~\cite{svre}, CWA~\cite{cwa}, SSA~\cite{ssa} and SIA~\cite{sia}. We report the CLIP$_{tar}$ score, which is the similarity between the response generated by the input images. (b) Comparison of image quality. We enlarge the local area of the adversarial examples to enhance visual effects. It is evident that adversarial examples generated by transfer-based attacks exhibit notable noise. Our method has better visual effects. Magnify images for improved contrast.} 
\label{fig:1}
\vskip -0.15in
\end{figure*}

In this paper, we propose AdvDiffVLM, an efficient framework that leverages diffusion models to generate natural, unrestricted, and targeted adversarial examples through score matching.
{Score matching, initially proposed by Hyvärinen et al.~\cite{score_ori},  is a computationally simple probability density estimation method.  It was later introduced into the field of image generation by Song et al.~\cite{score_gen}, demonstrating its ability to guide image generation toward specific target semantics by modifying the score function. Furthermore, Song et al.\cite{score} combined score matching with a diffusion model, significantly enhancing image quality. Inspired by these developments, we investigate the use of score matching to effectively and efficiently attack VLMs,  aiming to embed richer adversarial target semantics compared to existing methods like AdvDiffuser~\cite{advdiffuser}.}
{Specifically, we derive a score generation theory tailored for VLM attacks and propose the AEGE based on this theoretical foundation.}
Furthermore, to improve the naturalness of the outputs, we propose the GMGC module, which effectively distributes adversarial target semantics across the examples. This  prevents the concentration of adversarial features in specific regions, thereby improving overall image quality. In addition, we embed more target semantics into adversarial examples through multiple iterations, further enhancing the visual quality of the generated outputs.
As demonstrated in Figure~\ref{fig:1}, AdvDiffVLM  outperforms existing attack methods by generating targeted adversarial examples more efficiently while achieving superior transferability. Moreover, the generated adversarial examples exhibit enhanced naturalness, establishing AdvDiffVLM as a more effective tool for evaluating the adversarial robustness of VLMs.

We summarize our contributions as follows: 
\begin{itemize}
    \item {We explore existing adversarial attack techniques against VLMs and conduct research on more realistic and challenging scenarios, specifically focusing on targeted and transferable attacks.} Furthermore, we propose the AdvDiffVLM framework to efficiently generate targeted and transferable adversarial examples for VLMs. 
    \item {We present a score calculation method that embeds adversarial target semantics into the diffusion model, supported by theoretical analysis. 
    Additionally, we propose an adaptive ensemble method to better estimate the score. 
    Building on this theoretical foundation and adaptive ensemble method, we propose the AEGE that combines the generated gradient with score matching, embedding adversarial target semantics naturally in the inverse generation process of the diffusion model. } 
    \item {We apply an innovative use of GradCAM and propose the GCMG. In contrast to traditional applications, our method allows modifications to be made across the entire image while minimizing alterations to key areas, thus balancing attack capability with image quality.}
    \item  Extensive experiments show that our method generates targeted adversarial examples faster than SOTA adversarial attack methods in attacking VLMs, and the generated adversarial examples exhibit better transferability. 
    {In addition, our research identifies vulnerabilities in both open-source and commercial VLMs, offering insights toward developing more robust and trustworthy VLMs.}

\end{itemize}

\section{Related Work}\label{sec:related}

\subsection{Visual-Language Models (VLMs)}
Large language models (LLMs)~\cite{gpt3,t5,vicuna} have demonstrated great success in a variety of language-related tasks. The knowledge contained within these powerful LLMs has aided the development of VLMs. There are several strategies and models for bridging the gap between text and visual modalities~\cite{MLLM_survey,MLLM_survey2}. Some studies~\cite{flamingo,blip2} extract visual information from learned queries and combine it with LLMs to enhance image-based text generation. Models like LLaVA~\cite{llava} and MiniGPT-4~\cite{minigpt4} learn simple projection layers to align visual encoder features with LLM text embeddings. Some works~\cite{unidiffuser} train VLMs from scratch, which promotes better alignment of visual and textual modalities. In this paper, we focus on the adversarial robustness of these VLMs, with the goal of discovering security vulnerabilities and encouraging the development of more robust and trustworthy VLMs.

\subsection{Adversarial Attacks in VLMs} 
Adversarial attacks are classified as white-box or black-box attacks based on adversary knowledge, as well as targeted or untargeted attacks based on adversary objectives~\cite{attack_survey,attack_survey2,attack_survey3}. Studies have investigated the robustness of VLMs, focusing on adversarial challenges in visual question answering~\cite{vqa-attack} and image captioning~\cite{image-caption-attack}. However, most studies focus on traditional CNN-RNN-based models, which make assumptions about white-box access or untargeted goals, limiting their applicability in real-world scenarios. Recently, AttackVLM~\cite{attackvlm} implemented both transfer-based and query-based attacks on large open-source VLMs with black-box access and targeted goals. Nonetheless, this method is time-consuming due to its reliance on numerous VLM queries. In addition, \cite{attackbard} studied the adversarial robustness of VLMs using ensemble transfer-based attacks, assuming untargeted goals. In this paper, we investigate the adversarial robustness of VLMs against targeted transfer-based attacks. Initially, we evaluate VLM's robustness against current SOTA transfer-based attacks in conjunction with AttackVLM. We then examine the limitations of current methods and implement targeted improvements, culminating in the proposal of AdvDiffVLM. 



{Our method is most closely related to AttackVLM, as both aim to conduct adversarial attacks on VLMs. However, there are two notable differences.  First, while AttackVLM generates adversarial examples by estimating gradients through black-box model outputs, our approach leverages the transferability of adversarial examples to effectively attack multiple VLMs. This makes our method more versatile for attacking diverse VLMs. Furthermore, AttackVLM relies on extensive black-box queries to estimate gradients, making its generation process time-intensive. In contrast, our method utilizes the diffusion model's generation process, enabling faster creation of adversarial examples.}

{Second, in terms of methodology, AttackVLM combines gradient-based attacks with black-box query techniques, using the gradient-based component to initialize black-box queries effectively. In contrast, our approach adopts an unconstrained adversarial example generation framework grounded in generative models. By integrating adaptive gradient estimation with score matching, we embed the adversarial example generation process directly within the diffusion model’s workflow. To further enhance the quality of adversarial examples, we incorporate a GradCAM-guided masking technique, which refines the generated outputs.}


\subsection{Unrestricted Adversarial Examples}
Researchers are increasingly interested in unrestricted adversarial examples, as the $l_p$ norm distance fails to capture human perception \cite{uae1,uae2,advdiffuser,attention_sa,csfadv, colorfool}. 
Some approaches use generative methods to create unrestricted adversarial examples. For example, \cite{uae1} and \cite{uae2} {modify} the latent representation of GANs to produce unrestricted adversarial examples. However, due to the limited interpretability of GANs, the generated adversarial examples are of poor quality. Diffusion models \cite{ddpm} are SOTA generative models based on likelihood and theoretical foundations, sampling data distribution with high fidelity and diversity. AdvDiffuser~\cite{advdiffuser} {incorporates} the PGD~\cite{pgd} method into the diffusion model's reverse process, resulting in high-quality adversarial examples without restrictions. In this study, we explore using the diffusion model for generating unrestricted adversarial examples, focusing on modifying the score in the diffusion model's reverse process rather than adding noise to the latent image. We discuss the differences between our method and AdvDiffuser in Section~\ref{sec:difference}.

 
 
 
 

\section{Preliminaries}
\subsection{Diffusion Models}
In this work, we use diffusion models~\cite{ddpm, ddim, ldm} to generate unrestricted and targeted adversarial examples. In a nutshell, diffusion   models learn a denoising process from $x_T \sim \mathcal{N}\left(x_{T} ; 0, \mathbf{I}\right)$ to recover the data $x_0 \sim q(x_0)$ with a Markov chain and mainly include two processes: forward process and reverse process. Forward process defines a fixed Markov chain. Noise is gradually added to the image $x_0$ over $T$ time steps, producing a series of noisy images $\{x_1, x_2, \cdots, x_T\}$. Specifically, noise is added by $q\left(x_{t} \mid x_{0}\right):=\sqrt{\bar{\alpha}_{t}} x_{0}+\epsilon_1 \sqrt{1-\bar{\alpha}_{t}}, \epsilon_1 \sim \mathcal{N}(0,1)$, where $\alpha_{t}:=1-\beta_{t}$, $ \overline{\alpha_{t}}:=\prod_{s=1}^{t} \alpha_{s}$ and $\beta_t$ is a fixed variance to control the step sizes of the noise. The purpose of the reverse process is to gradually denoise from $x_T$ to obtain a series of $\{\tilde{x}_{T-1}, \tilde{x}_{T-2}, \cdots, \tilde{x}_1\}$, and finally restore $x_0$. It learns the denoising process through a denoising model $\varepsilon_{\theta}$, and the training  objective is $\mathcal{L}_{simple}:=E_{t \sim[1, T], \epsilon_1 \sim \mathcal{N}(0, \mathbf{I})}\left\|\varepsilon_{\theta}\left(x_{t}, t\right) - \epsilon_1 \right\|^{2}$.

\subsection{Problem Settings}
Then we give the problem setting of this paper.
We denote the victim VLM model as $f_{\xi}$, and aim to induce  $f_{\xi}$ to output the target response. This can be formalized as
\begin{equation}
\label{eq:1}
\begin{array}{r}
\begin{aligned}
\max\ &CS(g_{\psi }(f_{\xi }\left(\boldsymbol{x}_{\mathrm{adv}} ; \boldsymbol{c}_{\mathrm{in}}\right)),g_{\psi}(\boldsymbol{c}_{\mathrm{tar}})) \\
\text{s.t.} \ &D(\boldsymbol{x},\boldsymbol{x}_{\mathrm{adv} }) \le \epsilon
\end{aligned}
\end{array}
\end{equation}
where $\boldsymbol{x} \in \mathbb{R}^{3\times H \times W}$ represents the original image, $\boldsymbol{x}_\mathrm{adv}$ and  $\boldsymbol{c}_\mathrm{tar}$ respectively refer to adversarial example and adversarial target text, and $g_{\psi}(\cdot)$ denotes the CLIP text encoder. {Moreover}, $D(\boldsymbol{x},\boldsymbol{x}_{\mathrm{adv} }) \le \epsilon $ places a bound on a distance metric, and $CS(\cdot, \cdot)$ refers to the cosine similarity metric. Finally, $\boldsymbol{c}_\mathrm{in}$  {denotes} the input text.

Since $f_{\xi}$ is a black-box model, we generate adversarial examples on the surrogate model $\phi_{\psi }$ and transfer them to $f_{\xi}$. In addition, inspired by \cite{attackvlm}, matching image-image features can lead to better results, we define the problem as,
\begin{equation}
\label{eq:2}
\begin{array}{r}
\begin{aligned}
\begin{aligned}
\max\ &CS(\phi_{\psi }(\boldsymbol{x}_{\mathrm{adv}}),\phi_{\psi}(\boldsymbol{x}_{\mathrm{tar}})) \\
\text{s.t.} \ &D(\boldsymbol{x},\boldsymbol{x}_{\mathrm{adv} }) \le \epsilon
\end{aligned} 
\end{aligned}
\end{array}
\end{equation}
where $\boldsymbol{x}_\mathrm{tar}$ represents the target image generated by $\boldsymbol{c}_{\mathrm{tar}}$. We use stable diffusion~\cite{ldm} to implement the text-to-image generation. $\phi_{\psi }$ refers to CLIP image encoder.
Our study is the most realistic and challenging attack scenarios, i.e.,  targeted and transfer scenarios.

\begin{figure}[t]
\centering\includegraphics[width=0.9\linewidth]{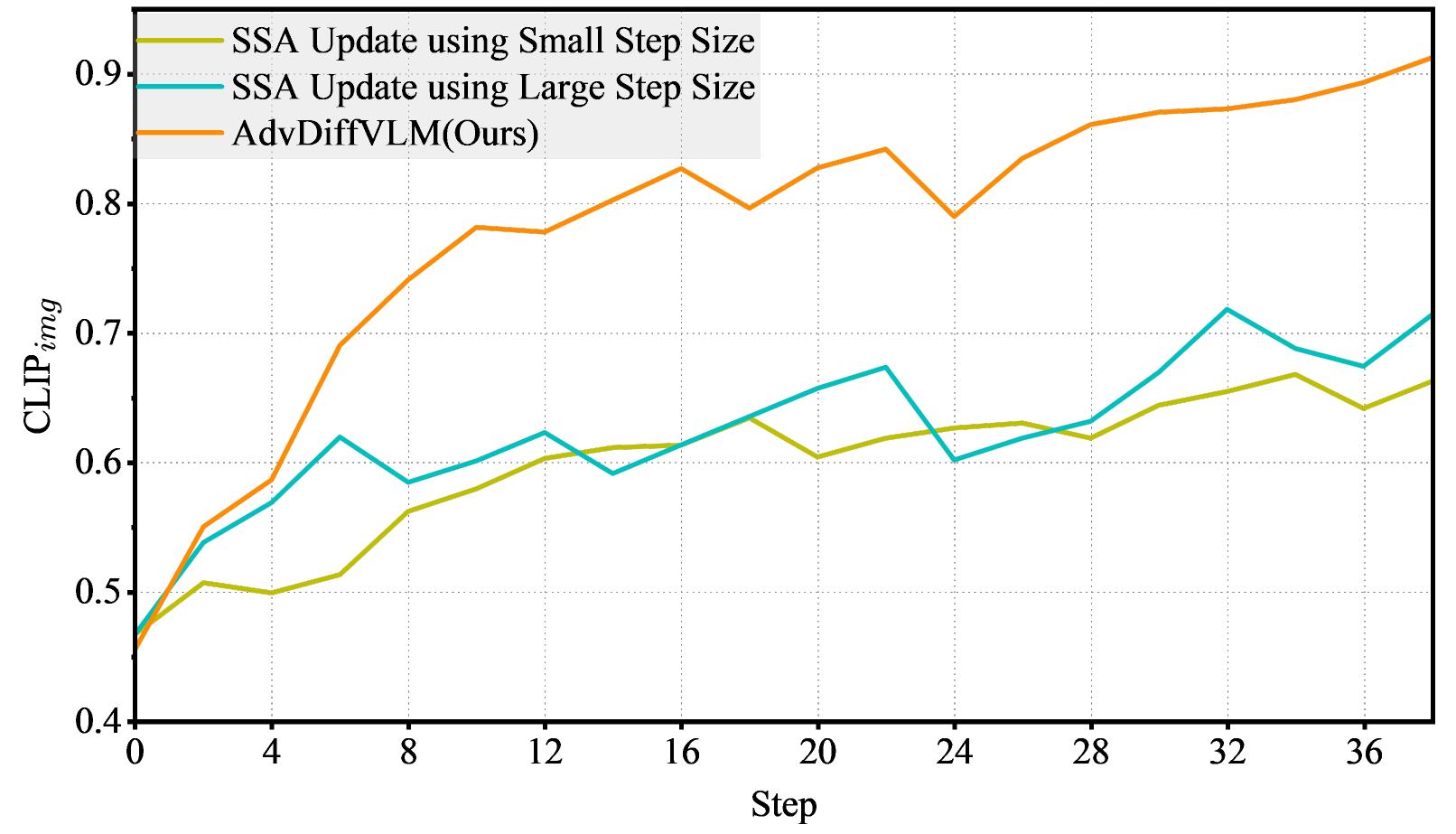}
\caption{The CLIP$_{img}$ score varies with the step sizes. Here, CLIP$_{img}$ is the similarity between the adversarial examples and the adversarial target images, which is calculated by the visual encoder of CLIP ViT-B/32. We choose SSA~\cite{ssa} as the representative of transfer-based attacks. } 
\label{fig:2}
\vspace{-4mm}
\end{figure}

\subsection{Rethinking Transfer-based Attacks}
\label{sec:transfer_attack}
Transfer-based attacks can effectively solve Eq.\ref{eq:2}. In this context, we assess the robustness of VLMs against current SOTA transfer-based attacks, in conjunction with AttackVLM. 
Specifically, we consider ensemble methods {including Ens~\cite{mifgsm}, SVRE~\cite{svre} and CWA~\cite{cwa}}, data augmentation methods {including SSA~\cite{ssa} and SIA~{\cite{sia}}}, and combinations of both{\footnote{{\url{https://github.com/xiaosen-wang/SIT} for SIA and \url{https://github.com/thu-ml/Attack-Bard} for others. we adapt the targeted attacks to untargeted attacks to better align with our scenario. Additionally, to ensure a fair comparison with our model, we modify the surrogate models by ensembling various CLIP visual encoders, including ResNet-50, ResNet-101, ViT-B/16, and ViT-B/32. Finally, we modify the loss function to cosine similarity loss to make it consistent with AttackVLM.}}}. We primarily employ the simple ensemble version of data augmentation attacks, as relying on a single surrogate model tends to yield poor performance. 
For hyperparameter settings, in all attacks, the value range of adversarial example pixels is [0,1]. We set the perturbation budget as $\epsilon = 16/255$ under the $\ell_{\infty}$ norm. The number of iterations $N_I$ for all attacks is set to 300. In addition, we use the MI-FGSM~\cite{mifgsm} method and set $\mu = 1$. {Furthermore, for SVRE, internal step size $\beta_{inter}=16/255/10$ and internal decay factor $\mu_2 = 1$. For CWA, CSE step size $\beta_{cse}=16/255/15$ and inner gradient ascent step $r=250$. For SSA, tuning factor $\rho = 0.5$, the number of spectrum transformations $N_t = 20$ and standard deviation $\sigma_s = 16 / 255$. For SSA, the number of the block $s=3$ and the number of image for gradient calculation $N_{grad} = 20$.}

The outcomes of these transfer-based attacks on VLMs are depicted in Figure~\ref{fig:1}. As illustrated, current transfer-based attacks face challenges such as slow adversarial example generation, noticeable noise within these examples, and limited transferability. The limitations of existing transfer-based attacks on VLMs are analyzed as follows: First, existing SOTA transfer-based attacks only access the original image during the optimization of Eq.\ref{eq:2}. Consequently, they employ small steps and strategies like data augmentation to tentatively approach the optimal solution, necessitating numerous iterations and resulting in high attack costs. As shown in Figure~\ref{fig:2}, using a larger step size results in pronounced fluctuations during the optimization process. This issue may be mitigated by leveraging score, which provides insights into the data distribution. By offering score guidance towards solving Eq.\ref{eq:2}, {faster} convergence is expected. Therefore score information can be considered in the design of new improved attack method. Second, existing transfer-based attacks introduce unnatural adversarial noises with limited transferability. Unrestricted adversarial examples can introduce more natural adversarial targeted semantics, increasing transferability. These imply that new transfer-based targeted attacks can consider unrestricted adversarial attacks.

\begin{figure*}[t]
\centering\includegraphics[width=\linewidth]{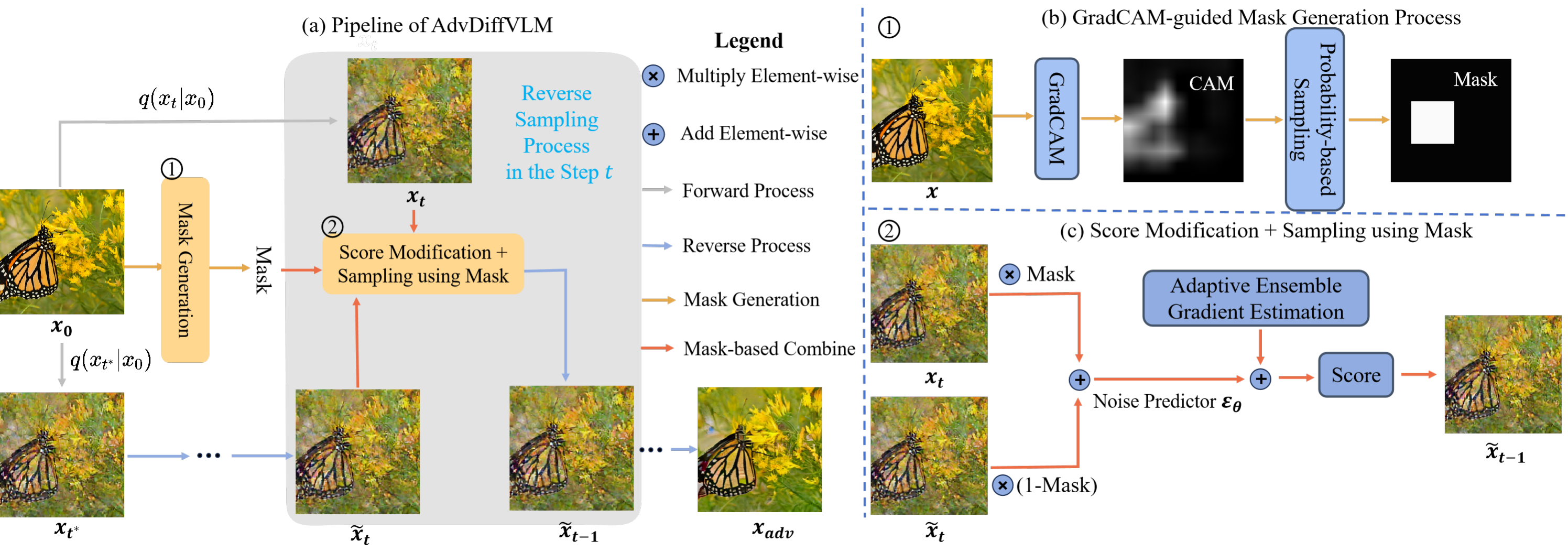}
\vskip -0.05in
\caption{ The main framework of the AdvDiffVLM for efficiently generating transferable unrestricted adversarial examples. AdvDiffVLM mainly includes two components: AEGE and GCMG. Details are respectively described in Secs.~\ref{sec:adaptive} and \ref{sec:gradcam}. Please refer to Section~\ref{sec:method} for specific symbol meanings.} 
\label{fig:3}
\vskip -0.15in
\end{figure*}

\section{Methodology}
\label{sec:method}

The main framework of AdvDiffVLM is illustrated in Figure~\ref{fig:3}. {First, we input the original image, $\boldsymbol{x}$, followed by the forward process $x_{t^{*}} \sim q\left(x_{t^{*}} \mid x_{0}\right)$ to obtain a noisy image, $x_{t^{*}}$. Subsequently, we apply the reverse denoising process. At each step, we first obtain the mask $\mathrm{m}$ from the GCMG module. This mask is then used to fuse the original noisy image, $x_{t}$, with the adversarial noisy image, $\tilde{x}_{t}$. Next, we apply the AEGE module to obtain gradient information, which we then use to calculate the score. Finally, we derive the next step of the noisy image based on the score matching method.}

In the following, we begin by presenting the motivation behind our approach and a theoretical analysis of our method. This is followed by a comprehensive explanation of the proposed AdvDiffVLM framework. Lastly, we highlight the key distinctions between our method and AdvDiffuser, emphasizing their unique features and contributions.

\subsection{{Motivation} and Theoretical Analysis}
{With the growing deployment of VLMs in critical applications such as autonomous driving and content moderation, ensuring their robustness against adversarial attacks has become essential for maintaining system security and reliability. While existing approaches have made notable progress in evaluating VLM robustness, they still face fundamental limitations in terms of efficiency and effectiveness. High computational overhead and limited transferability hinder the ability to comprehensively assess robustness across diverse VLMs. This challenge motivates our work to develop an efficient, high-quality, and transferable method for generating adversarial examples, thereby facilitating a more effective evaluation of VLM robustness. We achieve this by leveraging insights from diffusion models and score matching techniques.}

Specifically, we  focus on modeling adversarial attacks from a generative perspective, considering how to utilize the data distribution (score) of the generative model to produce natural, unrestricted and targeted adversarial examples. Additionally, as indicated in~\cite{unified_diffusion}, learning to model the score function is equivalent to modeling the negative of the noise, suggesting that score matching and denoising are equivalent processes. Thus, our method derives from integrating diffusion models and score matching, positioning it as a novel approach for generating high-quality, unrestricted, transferable and targeted adversarial examples.

Formally, we want to obtain distribution meeting the condition that the adversarial example has target semantic information during the reverse generation process 
\begin{equation}
p(x_{t-1}|x_t,f_{\xi }\left(\boldsymbol{x}_{\mathrm{adv}} ;\boldsymbol{c}_{\mathrm{in}}\right)=\boldsymbol{c}_{\mathrm{tar}})
\label{eq:3}
\end{equation}
where $x_t$ represents the latent image of the diffusion model. Next, we start from the perspective of score matching~\cite{score} and consider the score $\nabla \log p(x_{t-1}|x_t,\boldsymbol{c}_{\mathrm{tar}})$ of this distribution, where $\nabla$ is the abbreviation for $\nabla_{x_t}$. According to Bayes theorem,
\begin{equation}
\label{eq:4}
\begin{array}{l}\nabla \log p\left(x_{t-1} \mid x_{t}, \boldsymbol{c}_{\mathrm{tar}}\right)=\nabla \log \left(\frac{p\left(\boldsymbol{c}_{\mathrm{tar}} \mid x_{t-1}, x_{t}\right) \cdot p\left(x_{t-1} \mid x_{t}\right)}{p\left(\boldsymbol{c}_{\mathrm{tar}} \mid x_{t}\right)}\right) \\=\nabla \log p\left(\boldsymbol{c}_{\mathrm{tar}} \mid x_{t-1}, x_{t}\right)+\nabla \log p\left(x_{t-1} \mid x_{t}\right) \\\quad-\nabla \log p\left(\boldsymbol{c}_{\mathrm{tar}} \mid x_{t}\right) \\=\nabla \log p\left(\boldsymbol{c}_{\mathrm{tar}} \mid x_{t-1}\right)+\nabla \log p\left(x_{t} \mid x_{t-1}, \boldsymbol{c}_{\mathrm{tar}}\right) \\ -\nabla \log p\left(x_{t} \mid x_{t-1}\right) +\nabla \log p\left(x_{t-1} \mid x_{t}\right)-\nabla \log p\left(\boldsymbol{c}_{\mathrm{tar}} \mid x_{t}\right)\\=\nabla \log p\left(x_{t} \mid x_{t-1}, \boldsymbol{c}_{\mathrm{tar}}\right)-\nabla \log p\left(x_{t} \mid x_{t-1}\right) \\\quad+\nabla \log p\left(x_{t-1} \mid x_{t}\right)-\nabla \log p\left(\boldsymbol{c}_{\mathrm{tar}} \mid x_{t}\right)\end{array}
\end{equation}
$p\left(x_{t} \mid x_{t-1}, c_{tar}\right)$ and $p\left(x_{t} \mid x_{t-1}\right)$ respectively denote the add noise process with target text and the add noise process devoid of target semantics. From an intuitive standpoint, whether target text is present or not, the forward noise addition process follows a Gaussian distribution, and the added noise remains consistent, indicating that the gradient solely depends on $x_t$. The difference between $x_t$ without target text and $x_t$ with target text is minimal, as constraints are employed to ensure minimal variation of the adversarial {example} from the original image. Therefore, $\nabla \log p\left(x_{t} \mid x_{t-1}, \boldsymbol{c}_{\mathrm{tar}}\right)$ and $\nabla \log p\left(x_{t} \mid x_{t -1}\right)$ are approximately equal. So the final score is
$ \nabla \log p\left(x_{t-1} \mid x_{t}\right)-\nabla \log p\left(\boldsymbol{c}_{\mathrm{tar}} \mid x_{t}\right)$.

 Because score matching and denoising are equivalent processes, that is, $\nabla \log p\left(x_{t}\right)=-\frac{1}{\sqrt{1-\bar{\alpha}_{t}}} \varepsilon_{\theta}(x_t)$. Therefore we can get $\text{score~}$ $(\nabla \log p(x_{t-1}|x_t,\boldsymbol{c}_{\mathrm{tar}}))$, 
 \begin{equation}
     \label{eq:5}
     \text{score} = -(\frac{\varepsilon_{\theta}\left(x_{t}\right)}{\sqrt{1-\bar{\alpha}_{t}}} + \nabla \log p_{f_{\xi}}\left(c_{\operatorname{tar}} \mid x_{t}\right))
 \end{equation}
 where $\varepsilon_{\theta}$ is denoising model, and $\bar{\alpha}_t$ is the  hyperparameter.

Eq.\ref{eq:5} demonstrates that the score of $p(x_{t-1}|x_t,\boldsymbol{c}_{\mathrm{tar}})$ can be derived by incorporating gradient information into the inverse process of the diffusion model. Consequently, adversarial semantics can be incrementally embedded into adversarial examples based on the principle of score matching.

\subsection{Adaptive Ensemble Gradient Estimation (AEGE)}
\label{sec:adaptive}
Since $f_{\xi}$ is a black-box model and cannot obtain gradient information, we use surrogate model to estimate $\nabla \log p_{f_{\xi}}\left(c_{\operatorname{tar}} \mid x_{t}\right)$.
 As a scalable method for learning joint representations between text and images, CLIP~\cite{clip} can leverage pre-trained CLIP models to establish a bridge between images and text. Therefore we use the CLIP model as the surrogate model to estimate the gradient.

Specifically, we first add noise to the original image $\boldsymbol{x}$ by $t^{*}$ steps through the forward process $q(x_{t^{*}}|x_0)$ to obtain $x_{t^*}$, where $x_0 = \boldsymbol{x}$. Then, at each step of reverse process, we change score:
\begin{equation}
\label{eq:6}
\begin{array}{r}
\text{score}\! =\! \!-\!(\frac{1}{\sqrt{1-\bar{\alpha}_{t}}}\varepsilon_{\theta}\left(\tilde{x}_{t}\right)\! + \!s \nabla_{\tilde{x}_{t}} (CS(\phi_{\psi}(\tilde{x}_{t}),\phi_{\psi}(\boldsymbol{x}_\mathrm{tar}))))
\end{array}
\end{equation}
where $s$ is the adversarial gradient scale used to control the degree of score change and $\tilde{x}_{t}$ is the latent image in the reverse process.

We find that gradient estimation using only a single surrogate model is inaccurate. Therefore, we consider using a set of surrogate models  $\left\{\phi^{i}_{\Psi}\right\}_{i=1}^{N_m}$ to better estimate the gradient. Specifically, we make the following improvements to Eq.~\ref{eq:6}:
\begin{equation}
\label{eq:7}
\begin{array}{r}
\text{score} \!=\! \!-\!(\frac{\varepsilon_{\theta}\left(\tilde{x}_{t}\right)}{\sqrt{1-\bar{\alpha}_{t}}} \!+\! s \nabla_{\tilde{x}_{t}} (w_i \sum_{i=1}^{N_m}  CS(\phi_{\psi}^{i}(\tilde{x}_{t}),\phi_{\psi}^{i}(\boldsymbol{x}_\mathrm{tar}))))
\end{array}
\end{equation}
where $\mathbf{w} = (w_1, w_2,\cdots, w_{N_m}) $ represents the weight of cosine loss of different models.

Since different images have different sensitivities to surrogate models, only using simple ensemble cannot obtain optimal solution. Inspired by \cite{ensemble}, we propose a new adaptive ensemble method, and obtain $\mathbf{w}$ in Eq.~\ref{eq:7} in the following way:
\begin{equation}
\label{eq:8}
\begin{array}{r}
w_i(t) = \displaystyle\frac{\textstyle \sum_{j=1}^{N_m}\exp(\tau \mathcal{L}_j(t+1)/\mathcal{L}_j(t+2))}{ {N_m\exp(\tau \mathcal{L}_i(t+1)/\mathcal{L}_i(t+2))} } 
\end{array}
\end{equation}
where $\tau$ refers to the temperature. A larger $\tau$ makes all weights close to 1. $\mathcal{L}_i = CS(\phi_{\psi}^{i}(\tilde{x}_{t}),\phi_{\psi}^{i}(\boldsymbol{x}_\mathrm{tar}))$. We initialize $\{w_i(t^*)\}_{i=1}^{N_m}$ and $\{w_i(t^*-1)\}_{i=1}^{N_m}$ to 1. Through Eq.~\ref{eq:8}, we reduce the weight of surrogate models with fast-changing losses to ensure that gradient estimations of different surrogate models are updated simultaneously.

\begin{figure}[t]
\centering\includegraphics[width=\linewidth]{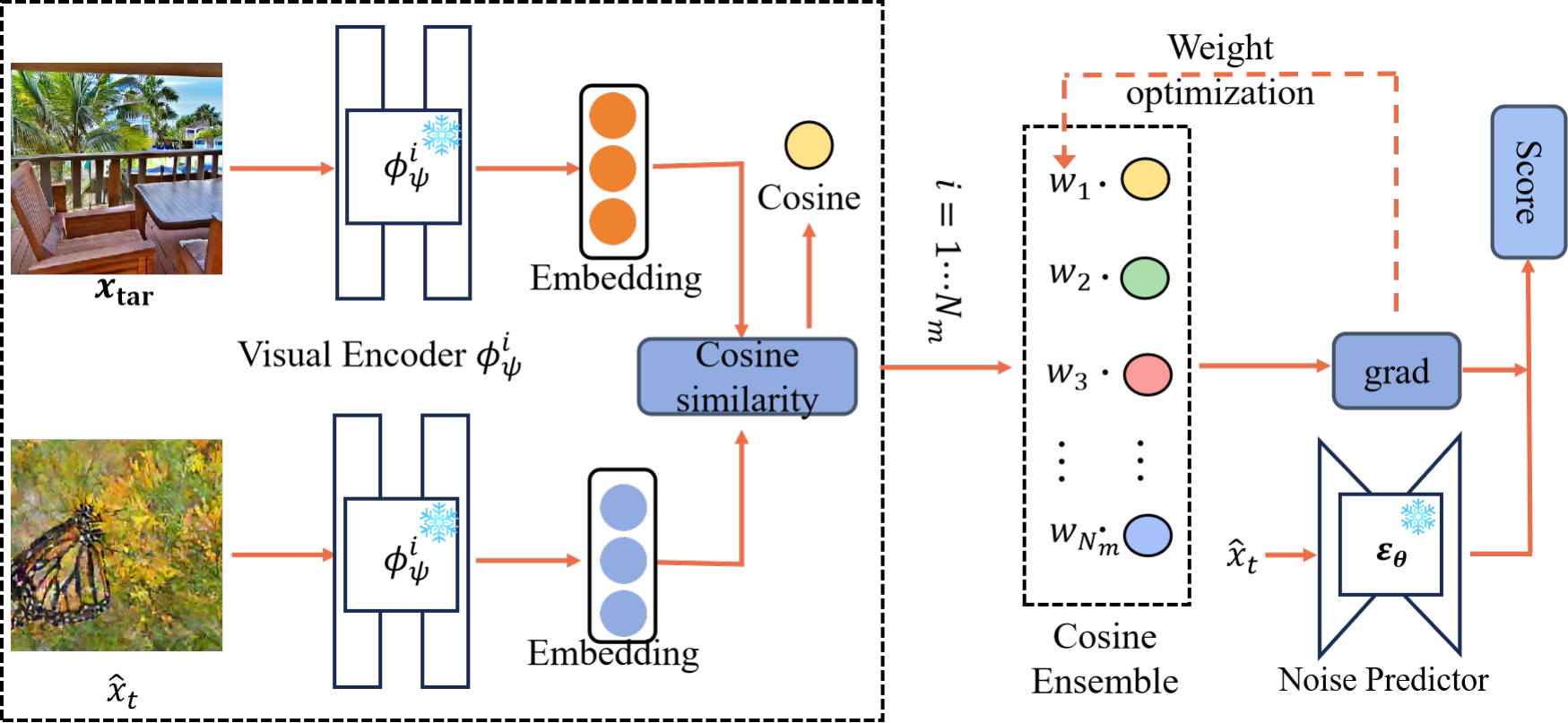}
\vskip -0.05in
\caption{ {The pipeline of the AEGE.}} 
\label{fig:aege}
\vskip -0.05in
\end{figure}

{Figure~\ref{fig:aege} presents the detailed visualization results of AEGE. Specifically, both the target image and the current adversarial example are independently input into $N_m$ visual encoders to obtain $N_m$ cosine similarity values. Here, $\hat{x}_t$ represents the current adversarial example derived from the mask, as described in Sec.~\ref{sec:gradcam}. The gradient, denoted as grad, is then computed by applying weights $\mathbf{w}$ to these values. Finally, this weighted result is combined with the noise prediction value to obtain the final score.}

Finally, we set the perturbation threshold $\delta$, and then clip the adversarial gradient to  ensure the naturalness of the synthesized adversarial examples. 

\subsection{GradCAM-guided Mask Generation (GCMG)}
\label{sec:gradcam}

We detailed AEGE earlier but observed that relying solely on it generates obvious adversarial features in specific areas, leading to poor visual effects. To balance visual quality and attack capabilities, we propose GCMG, which uses a mask to combine the forward noisy image $x_t$ and the generated image $\tilde{x}_{t}$. This combination distributes adversarial semantics across the image, enhancing the natural visual quality of adversarial examples.

First, we utilize GradCAM~\cite{gardcam} to derive the class activation map $CAM$ of $\boldsymbol{x}$ with respect to ground-truth label $\boldsymbol{y}$. $CAM$ assists in identifying important and non-important areas in the image. Subsequently, we clip the $CAM$ values to the range $[0.3,0.7]$ and normalize them to obtain the probability matrix $\mathrm {P}$. 
We sample according to the $\mathrm {P}$ to obtain the coordinate $(\mathrm {x,y})$, and then set the $\mathrm {k} \times \mathrm {k}$ area around $(\mathrm {x,y})$ to be 1 and remain other areas to obtain mask $\mathrm{m}$. Here, $\mathrm{m}$ has the same shape as $\tilde{x}_{t}$. This approach disperses more adversarial features in non-important areas and less in important areas of adversarial examples, improving the natural visual effect of adversarial examples.

At each step $t$, we combine $x_t$ and $\tilde{x}_{t}$ as following:
\begin{equation}
\label{eq:9}
\hat{x}_{t} = \mathrm{m}\odot x_t+(1-\mathrm{m}) \odot \tilde{x}_t
\end{equation}
where {$\mathrm{m}$ denotes the final mask,} $\odot$ refers to Hadamard Product. Afterwards, we can obtain new $\text{score}$ by integrating $\varepsilon_{\theta}\left(\hat{x}_{t}\right)$ with the estimated gradient and then use $\tilde{x}_{t-1} = - \sqrt{1-\bar{\alpha}_{t}} \times \text{score}$ for sampling.

Finally, we take the generated adversarial example as $x_0$, and iterate $N$ times to embed more target semantics into it. We provide a complete algorithmic overview of AdvDiffVLM in Algorithm~\ref{alg:1}.

\begin{algorithm}[t]
   \caption{The overall algorithm of AdvDiffVLM}
   \label{alg:1}
   \footnotesize{\KwIn{Original image $\boldsymbol{x}$, $N_m$ surrogate models $\phi^{i}_{\theta}$, adversarial guidance scale $s$, reverse generation process timestep $t^*$, mask area size $\mathrm{k}$, perturbation threshold $\delta$, temperature $\tau$, adversarial target image $\boldsymbol{x}_\mathrm{tar}$, Number of iterations $N$.}}
   \footnotesize{\KwOut{adversarial example $\boldsymbol{x}_\mathrm{adv}$.}}

   Initialize $\{w_i\}_{i=1}^{N_m} = 1$, $CAM$, $x_0=\boldsymbol{x}$\;
   Sample $x_{t^*}\sim q(x_{t^*}|x_0)$, let $\tilde{x}_{t^*} = \bar{x}_{t^*} = x_{t^*}$\;
   \For{$n\leftarrow 1, \cdots, N$}{
   \For{$t\leftarrow t^*, \cdots, 1$}{
        Get mask $\mathrm{m}$ according to $CAM$ \;
         \rightline{\tcp{{Mask generation;}}}
        $x_t\sim q(x_t|x_0)$\; 
        $\hat{x}_{t} = \mathrm{m}\odot x_t+(1-\mathrm{m}) \odot \tilde{x}_t$\; \rightline{\tcp{{Mask-based combination;}}}
        $w_i = \displaystyle\frac{\textstyle \sum_{j=1}^{N_m}\exp(\tau \mathcal{L}_j(t+1)/\mathcal{L}_j(t+2))}{ {N_m\exp(\tau \mathcal{L}_i(t+1)/\mathcal{L}_i(t+2))} }$\; \rightline{\tcp{{Weight optimization;}}}
        $g = \nabla_{\hat{x}_{t}} (w_i \sum_{i=1}^{N_m}  CS(\phi_{\psi}^{i}(\hat{x}_{t}),\phi_{\psi}^{i}(\boldsymbol{x}_\mathrm{tar})))$\; \rightline{\tcp{{Ensemble gradient estimation;}}}
         $g = clip(g, -\delta, \delta)$\; 
         $\text{score} =\varepsilon_{\theta}\left(\hat{x}_{t}\right) / \sqrt{1-\bar{\alpha}_{t}} + s \cdot g$\; \rightline{\tcp{{Score calculation;}}}
         $\tilde{x}_{t-1} = {(\hat{x}_t+(1-\alpha_t)\cdot \text{score})}/{\sqrt{\alpha_t}}$\; \rightline{\tcp{{Score matching;}}}
   }
   }
   Return $\boldsymbol{x}_\mathrm{adv} = \tilde{x}_0$
\end{algorithm}

\subsection{Differences from AdvDiffuser}
\label{sec:difference}
Both our method and AdvDiffuser~\cite{advdiffuser} produce unrestricted adversarial examples using the diffusion model. Here, we discuss the distinctions between them, highlighting our contributions.

\textbf{Tasks of varying difficulty levels:} AdvDiffuser is oriented towards classification models, while our research targets the more intricate Vision-Language Models (VLMs). Initially, within the realm of classification tasks, each image is associated with a singular label. Conversely, in the image-to-text tasks, images may be linked to numerous text descriptions. When faced with an attack targeting a singular description, VLMs have the capability to generate an alternate description, thereby neutralizing the attack's effectiveness. As a result, our task presents a greater challenge.

\begin{figure}[t]
\centering\includegraphics[width=0.9\linewidth]{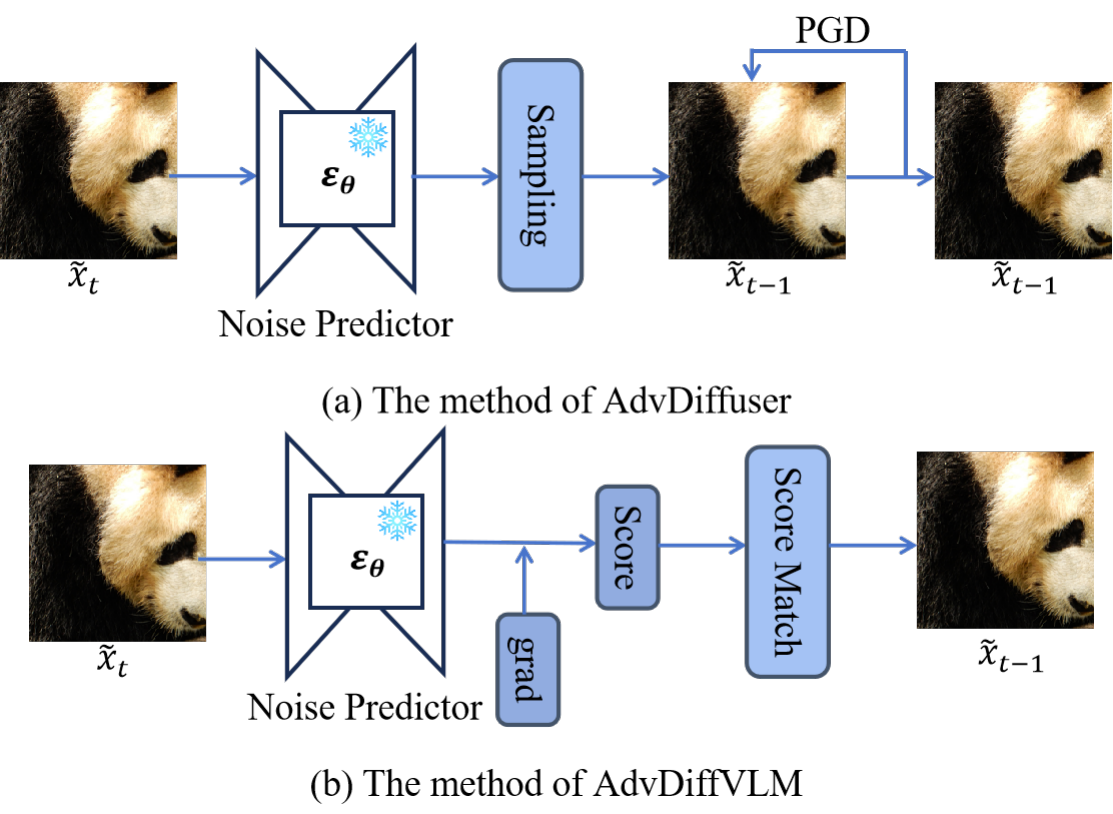}
\vskip -0.05in
\caption{ {Different theoretical foundations and implementation methods between AdvDiffuser and our method. Where ``Sampling" refers to $\tilde{x}_{t-1} = \left( \tilde{x}_t - {(1-\alpha_t)} \cdot  \boldsymbol{\varepsilon}_\theta (\tilde{x}_t, t) / {\sqrt{1-\bar{\alpha}_t}} \right) / {\sqrt{\alpha_t}} $ and ``Score Match" refers to $\tilde{x}_{t-1} = {(\tilde{x}_t+(1-\alpha_t)\cdot \text{score})}$}} 
\label{fig:diff}
\vskip -0.1in
\end{figure}

{\textbf{Different theoretical foundations and implementation methods:} AdvDiffuser utilizes PGD~\cite{pgd} to introduce high-frequency adversarial noise, while our method employs score matching to incorporate target semantics. These theoretical distinctions lead to differences in implementation: Without considering the mask, AdvDiffuser operates on the latent image $\tilde{x}_t$, adding adversarial noise directly to $\tilde{x}_t$. In contrast, our method modifies the predicted noise $\varepsilon_{\theta}\left(\tilde{x}_{t}\right)$, obtaining a score that encodes adversarial target semantics. For clarity, we include a visual comparison, as shown in Figure~\ref{fig:diff}.}
Furthermore, our approach obviates the need for initiating with Gaussian noise, initially introducing noise to $\boldsymbol{x}$ through $t^*$ steps, followed by the application of adversarial gradient to modify score, thereby facilitating more efficient generation of adversarial examples. 

\textbf{Distinct schemes of GradCAM utilization:} The GradCAM mask utilized by AdvDiffuser leads to restricted modification of crucial image areas, rendering it inadequate for  image-based attacks. Addressing this issue, we introduc the GCMG. Contrary to utilizing GradCAM results directly as a mask, we employ them as a directive to generate the mask further. This not only guarantees a likelihood of modification across all image areas but also secures minimal alteration of significant areas, striking a balance between image quality and attack ability. 

\begin{table}[t]
  \centering
    \caption{The details of victim VLMs, include code and configuration.}
    \renewcommand\arraystretch{1.35}
\resizebox{0.48\textwidth}{!}{
\begin{tabular}{c|c|c} 
    \toprule
 \textbf{Models}  & \textbf{Code} & \textbf{Version}\\
\midrule
   Unidifusser & https://github.com/thu-ml/unidiffuser & / \\


BLIP2 & https://github.com/salesforce/LAVIS & (blip2\_opt, pretrain\_opt2.7b)\\

MiniGPT-4 & https://github.com/Vision-CAIR/MiniGPT-4 & (Vicuna 7B)\\

LLaVA & https://github.com/haotian-liu/LLaVA & (Vicuna, llava-v1.5-7b)\\
\
Img2LLM & https://github.com/salesforce/LAVIS & (img2prompt\_vqa, base)\\
\bottomrule
    \end{tabular}}
    \label{tab:3}
 \vskip -0.05in    
\end{table}%

\begin{table*}[t]
\vskip -0.05in
    \centering
    \caption{Comparison with existing SOTA attack methods, where the best result is \textbf{bolded}. {We also report the standard deviation of the results. } Note that we use four versions of the CLIP visual encoder, including Resnet50, Resnet101, ViT-B/16 and ViT-B/32, as surrogate models.  Since Unidiffuser uses ViT-B/32 as the visual encoder, it is a gray box scenario, which we indicate with *. {In addition}, we provide the average time (s) for each strategy to craft
a single $x_{adv}$. The \colorbox{gray!20}{shaded parts} represent our proposed method.}
    \vskip -0.05in
\resizebox{\textwidth}{!}{
    \begin{tabular}{l|cc|cc|cc|cc|cc|c}
    \toprule
    & \multicolumn{2}{c|}{Unidiffuser*}
    & \multicolumn{2}{c|}{BLIP2}
    & \multicolumn{2}{c|}{MiniGPT-4}
    & \multicolumn{2}{c|}{LLaVA}
    & \multicolumn{2}{c|}{Img2LLM}
    & \multirow{2}{*}{Time(s)}\\
    & CLIP$_{tar}$ $\uparrow$ & ASR $\uparrow$ 
    & CLIP$_{tar}$ $\uparrow$ & ASR $\uparrow$
    & CLIP$_{tar}$ $\uparrow$ & ASR $\uparrow$
    & CLIP$_{tar}$ $\uparrow$ & ASR $\uparrow$
    & CLIP$_{tar}$ $\uparrow$ & ASR $\uparrow$

    & 
    \\ \midrule\midrule
Original        &
0.4770$_{\pm 0.0017}$        & 0.0\%$_{\pm 0.00\%}$        &
0.4931$_{\pm 0.0027}$         & 0.0\%$_{\pm 0.00\%}$         & 0.4902$_{\pm 0.0030}$        & 0.0\%$_{\pm 0.00\%}$         &
0.5190$_{\pm 0.0052}$         & 0.0\%$_{\pm 0.00\%}$         & 0.5288$_{\pm 0.0039}$        & 0.0\%$_{\pm 0.00\% }$        &
/         
\\\midrule\midrule
Ens        &
0.7353$_{\pm 0.0012}$         & 99.1\%$_{\pm 0.14\%}$          &
0.5085$_{\pm 0.0019}$         & 0.9\%$_{\pm 0.11\%}$         & 0.4980$_{\pm 0.0035}$        & 1.8\%$_{\pm 0.14\%}$          &
0.5366$_{\pm 0.0052}$         & 3.5\%$_{\pm 0.20\%}$          & 0.5297$_{\pm 0.0027}$        & 4.5\%$_{\pm 0.22\%}$          &
69 \\
SVRE        &
0.7231$_{\pm 0.0020}$        & 100.0\%$_{\pm 0.00\%}$       &
0.5190$_{\pm 0.0023}$         & 2.4\%$_{\pm 0.15\%}$        & 0.5107$_{\pm 0.0029}$        & 2.2\%$_{\pm 0.12\%}$         &
0.5385 $_{\pm 0.0049}$        & 4.6\%$_{\pm 0.18\%}$        & 0.5292$_{\pm 0.0035}$        & 3.8\%$_{\pm 0.17\%}$         &
125 \\
CWA        &
0.7568$_{\pm 0.0016}$   & 100.0\%$_{\pm 0.00\%}$      &
0.5249$_{\pm 0.0022}$         & 5.2\%$_{\pm 0.27\%}$         & 0.5211$_{\pm 0.0033}$        & 3.8\%$_{\pm 0.20\%}$         &
0.5493$_{\pm 0.0057}$         & 7.1\%$_{\pm 0.26}\%$         & 0.5346$_{\pm 0.0042}$        & 5.4\%$_{\pm 0.04\%}$         &
101 \\\midrule
SSA-Ens        &
0.7275$_{\pm 0.0031}$         & 100.0\%$_{\pm 0.00\%}$        &
0.5539$_{\pm 0.0050}$         & 9.2\%$_{\pm 0.42\%}$        & 0.5175$_{\pm 0.0052}$        & 10.1\%$_{\pm 0.33\%}$         &
0.6098$_{\pm 0.0056}$         & 37.5\%$_{\pm 0.49\%}$         & 0.5629$_{\pm 0.0050}$        & 19.6\%$_{\pm 0.31\%}$         &
879 \\
SSA-SVRE        &
0.7217$_{\pm 0.0039}$        & 100.0\%$_{\pm 0.00\%}$        &
0.5776$_{\pm 0.0046}$         & 18.7\%$_{\pm 0.40\%}$         & 0.5395$_{\pm 0.0056}$        & 16.5\%$_{\pm 0.47\%}$         &
0.6005$_{\pm 0.0063}$         & 40.2\%$_{\pm 0.57\%}$         & 0.5625$_{\pm 0.0067}$        & 18.4\%$_{\pm 0.39\%}$        &
1012 \\
SSA-CWA       &
0.7485$_{\pm 0.0024}$          & 100.0\%$_{\pm 0.00\%}$       & 
0.5888$_{\pm 0.0041}$         & 23.3\%$_{\pm 0.57\%}$         & 0.5407$_{\pm 0.0057}$        & 20.6\%$_{\pm 0.45\%}$          &
0.6152$_{\pm 0.0070}$         & 40.7\%$_{\pm 0.52\%}$         & 0.5634$_{\pm 0.0061}$        & 20.4\%$_{\pm 0.33\%}$         &
1225 \\\midrule
SIA-Ens        &
0.7377$_{\pm 0.0058}$         & 100.0\%$_{\pm 0.00\%}$        &
0.5956$_{\pm 0.0074}$         & 49.6\%$_{\pm 1.40\%}$         & 0.5605$_{\pm 0.0064}$        & 40.4\%$_{\pm 1.06\%}$         &
0.7158$_{\pm 0.0085}$ & 84.7\%$_{\pm 1.80\%}$  & 0.6337$_{\pm 0.0073}$    & 27.0\%$_{\pm 1.27\%}$ &
483 \\
SIA-SVRE       &
0.7302$_{\pm 0.0066}$         & 100.0\%$_{\pm 0.00\%}$        &
0.6102$_{\pm 0.0068}$         & 50.1\%$_{\pm 0.79\%}$  & 0.5782$_{\pm 0.0080}$        & 46.4\%$_{\pm 1.17\%}$        &
0.7122$_{\pm 0.0079}$  & {88.3\%$_{\pm 1.73\%}$}   & 0.6305$_{\pm 0.0087}$        & {35.4\%$_{\pm 1.54\%}$}         &
596 \\
SIA-CWA        &
0.7498$_{\pm 0.0053}$ & 100.0\%$_{\pm 0.00\%}$      &
0.6135$_{\pm 0.0085}$         & {51.8\%$_{\pm 1.18\%}$}         & 0.5810$_{\pm 0.0064}$        & 47.8\%$_{\pm 1.24\%}$         &
0.7194$_{\pm 0.0086}$   & {89.5\%$_{\pm 2.95\%}$} & 0.6401$_{\pm 0.0078}$  & {40.6\%$_{\pm 1.30\%}$}  &
732 
\\\midrule\midrule
AdvDiffuser$_{ens}$        &
0.6774$_{\pm 0.0037}$         & 86.7\%$_{\pm 1.93\%}$              &
0.5396$_{\pm 0.0034}$         & 8.6\%$_{\pm 0.26\%}$         & 0.5371$_{\pm 0.0041}$        & 8.2\%$_{\pm 0.37\%}$         &
0.5507$_{\pm 0.0071}$         & 25.3\%$_{\pm 0.37\%}$         & 0.5395$_{\pm 0.0063}$        & 11.5\%$_{\pm 0.25\%}$         &
574 \\
AdvDiffuser$_{adaptive}$        &
0.6932$_{\pm 0.0029}$         & 88.9\%$_{\pm 1.75\%}$                 &
0.5424$_{\pm 0.0062}$         & 10.4\%$_{\pm 0.35\%}$         & 0.5391$_{\pm 0.0046}$        & 9.6\%$_{\pm 0.30\%}$         &
0.5595$_{\pm 0.0064}$         & 27.4\%$_{\pm 0.41\%}$         & 0.5502$_{\pm 0.0060}$        & 14.8\%$_{\pm 0.32\%}$         &
602 
\\\midrule\midrule

\rowcolor{gray!20}AdvDiffVLM        &
0.7502$_{\pm 0.0072}$         & 100.0\%$_{\pm 0.00\%}$                  &
\textbf{0.6435}$_{\pm 0.0101}$ & \textbf{66.7\%$_{\pm 1.86\%}$}  & \textbf{0.6145}$_{\pm 0.0096}$  & \textbf{58.6\%$_{\pm 2.07\%}$}   &
\textbf{0.7206}$_{\pm 0.0113}$        & \textbf{91.2\%$_{\pm 2.35\%}$}        & \textbf{0.6521}$_{\pm 0.0107}$ & \textbf{43.8\%$_{\pm 1.92\%}$}  &
139 \\
\bottomrule
    \end{tabular}}
    \label{tab:2}
 \vskip -0.05in
\end{table*}

\section{Experiments}
\subsection{Experimental Setup}
\label{sec:experiment_setup}
\textbf{Datasets and victim VLMs:} Following \cite{attackbard}, we use NeurIPS'17 adversarial competition dataset, compatible with ImageNet, for all the experiments. {In addition}, we select 1,000 text descriptions from the captions of the MS-COCO dataset as our adversarial target texts and then use Stable Diffusion~\cite{ldm} to generate 1,000 adversarial targeted images. For the victim VLMs,  SOTA open-source models are evaluated, including Unidiffuser~\cite{unidiffuser}, BLIP2~\cite{blip2},  MiniGPT-4~\cite{minigpt4}, LLaVA~\cite{llava} and Img2LLM~\cite{image2llm}. The details are shown in Table~\ref{tab:3}. 
Among them, Unidiffuser is a gray-box model, and the others are black-box models.

\textbf{Baselines:} We compare with  AdvDiffuser~\cite{advdiffuser} and other SOTA transfer-based attackers described in Section~\ref{sec:transfer_attack}.  Since AdvDiffuser is used for classification models, we use cosine similarity loss instead of classification loss for adversarial attacks on VLMs. 
For a fair comparison, we implement the ensemble version of AdvDiffuser, including simple ensemble and adaptive ensemble, which are denoted as AdvDiffuser$_{ens}$, AdvDiffuser$_{adaptive}$ respectively. For hyperparameters (in AdvDiffuser), we choose $T=200, \sigma=0.4, I=25$.

\textbf{Evaluation metrics:}  Following \cite{attackvlm}, we adopt CLIP score between the generated responses from victim models and predefined targeted texts, as computed by ViT-B/32 text encoder, refered as CLIP$_{tar}$. 
We adopt the method of calculating the attack success rate (ASR) in~\cite{attackbard}, positing that an attack is deemed successful solely if the image description includes the target semantic main object. 
In order to measure the quality of adversarial examples and the perceptibility of applied perturbations, we use four evaluation metrics: SSIM~\cite{ssim}, FID~\cite{fid}, LPIPS~\cite{lpips} and BRISQUE~\cite{brisque}.

\textbf{Implementation details:} Since our adversarial diffusion sampling does not require additional training to the original diffusion model, we use the pre-trained diffusion model in our experiment. We adapt  LDM~\cite{ldm} with DDIM sampler~\cite{ddim}  (using T = 200 diffusion steps) {and select the version trained on ImageNet}. Additionally, we use four versions of CLIP~\cite{clip}, namely ResNet-50, ResNet-101, ViT-B/16, and ViT-B/32, {each trained on 400 million unpublished image-text pairs.}
For other hyperparameters, we use $s=35, \delta =0.0025, t^*=0.2, \mathrm{k}=8, \tau =2$ and $N=10$. All the experiments are conducted on a Tesla A100 GPU with 40GB memory.

\begin{figure*}[htbp]
\centering\includegraphics[width=\linewidth]{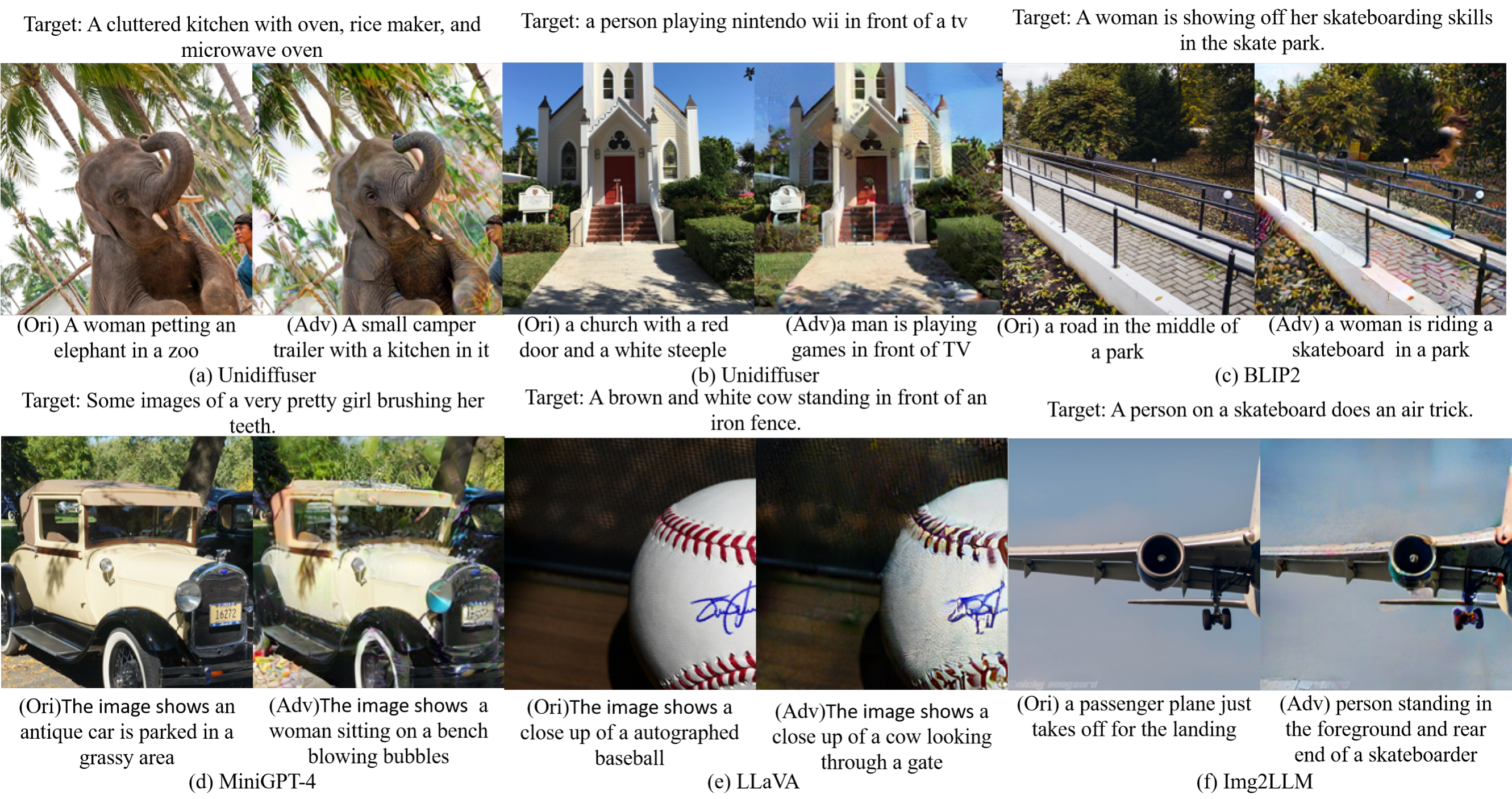}
\vskip -0.13in
\caption{Visualization of the attack results of our method on  various open-source VLMs. We show the adversarial target text above the image, and display the image caption results of original image and adversarial example below the image.} 
\label{fig:5}
\vskip -0.2in
\end{figure*}

\begin{table}
  \centering
    \caption{The result of attacking commercial VLMs.  We report ASR and provide the average time (s) for each strategy to craft a single $x_{adv}$. {The best result is \textbf{bolded}.}}. 
    \vskip -0.05in

\begin{tabular}{c|cccc|c} 
    \toprule
   & GPT-4V & Gemini  & Copilot & ERNIE Bot & Time(s)  \\
\midrule
 No attack & 0\% & 0\% & 0\% & 0\% &/ \\
 SIA-CWA & 35\% & 12\% & 25\% & 50\% & 732\\
 \rowcolor{gray!20}AdvdiffVLM & \textbf{37\%} & \textbf{17\%} & \textbf{26\%} & \textbf{58\%} & \textbf{139}\\
    \bottomrule
    \end{tabular}
    \label{tab:7}
    \vskip -0.15in
\end{table}%

\subsection{Main Experiments}
In this subsection, we evaluate the effectiveness of our method in targeted and transferable scenarios. Specifically, we first quantitatively compare the transferability of our approach against baseline methods on both open-source and commercial VLMs. Following this, we present qualitative results of our method applied to these VLMs. Lastly, we analyze the model's complexity and demonstrate its practical efficiency.

\label{sec:comparison_results}
\textbf{Quantitative results on open source VLMs.} To validate the effectiveness of AdvDiffVLM, we quantitatively evaluate the transferability of adversarial examples generated by AdvDiffVLM and baseline methods on various open source VLMs. As shown in Table~\ref{tab:2}, all methods demonstrate favorable attack results in gray box scenarios. In the transfer attack scenario, our method yields the best results. For example, on BLIP2, our method improves CLIP$_{tar}$ and ASR by 0.0200 and 10.9\% , respectively, when compared to SIA-CWA. Furthermore, our method generates adversarial examples much faster than baselines. Specifically, when compared to AdvDiffuser, SIA and SSA methods, our method generates adversarial examples 5x to 10x faster. Experimental results show that our method generates adversarial examples with better transferability at a faster rate, demonstrating its superiority. {To ensure statistical significance, we repeat each experiment three times and report the standard deviation.}

Additionally, it has been observed that AdvDiffuser exhibits suboptimal performance in challenging attack scenarios, particularly against VLMs. This is attributed to its direct application of GradCAM as the mask, which restricts the modifiable area for adversarial examples in demanding tasks, thereby diminishing attack effectiveness. Simultaneously, AdvDiffuser employs high-frequency adversarial noise to alter semantics. This adversarial noise, being inherently fragile, is significantly mitigated during the diffusion model's reverse process, further diminishing its attack potential on complex tasks. These observations validate the advantages of our GCMG and score matching idea.

\textbf{Quantitative results on commercial VLMs.} We conduct a quantitative evaluation of commercial VLMs such as OpenAI's GPT-4V\footnote{\url{https://chat.openai.com/}}, Google’s Gemini\footnote{\url{https://gemini.google.com/}}, Microsoft’s Copilot\footnote{\url{https://copilot.microsoft.com/}}, and Baidu’s ERNIE Bot\footnote{\url{https://yiyan.baidu.com/}}. We choose SIA-CWA to represent baselines and ASR as an evaluation metric. We {choose} 100 images from the NeurIPS'17 adversarial competition dataset and 100 text descriptions from the MS-COCO dataset as target texts. Table~\ref{tab:7} presents the experimental results. Our method outperforms SIA-CWA in terms of attack success rate, demonstrating its superior transferability.

\textbf{Qualitative results on open source VLMs.} We then present  visualizations depicting the outcomes of our method's attacks on open source VLMs, as illustrated in Figure~\ref{fig:5}. Considering the image caption task, we focus on two models: Unidiffuser and BLIP2. Considering the VQA task, we focus on MiniGPT-4, LLaVA and Img2LLM. In the case of MiniGPT-4, the input text is configured as ``What is the image showing?". For LLaVA, the input text is set to ``What is the main contain of this image?", and the prefix ``The main contain is" is omitted in the output. For Img2LLM, the input text is configured as ``What is the content of this image?". Our method demonstrates the capability to effectively induce both gray-box and black-box VLMs to produce adversarial target semantics. For example, in the case of LLaVA's attack, we define the adversarial target text as ``A cake that has various gelatins in it." LLaVA generate the response ``The main contain is a close-up view of a partially eaten cake with chocolate and white frosting." as the target output, while the original image's content is described as ``The main contain is a bird, specifically a seagull, walking on the beach near the water.".

\begin{figure}[t]
\centering\includegraphics[width=\linewidth]{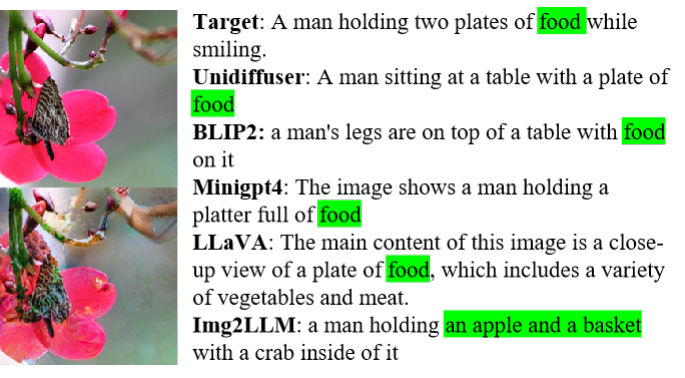}
\vskip -0.13in
\caption{{Visualization results of using the same adversarial example to attack various victim models. Where the top one is the original image, and the bottom one is the adversarial example. The target semantics is marked in green.}} 
\label{fig:same}
\vskip -0.2in
\end{figure}

{In addition, we visualize the outputs of various victim models with the same adversarial example, as shown in Figure~\ref{fig:same}. The visualization demonstrates that the adversarial example successfully induces all victim models to produce the target semantics.}

\begin{figure*}[htbp]
\centering\includegraphics[width=0.85\linewidth]{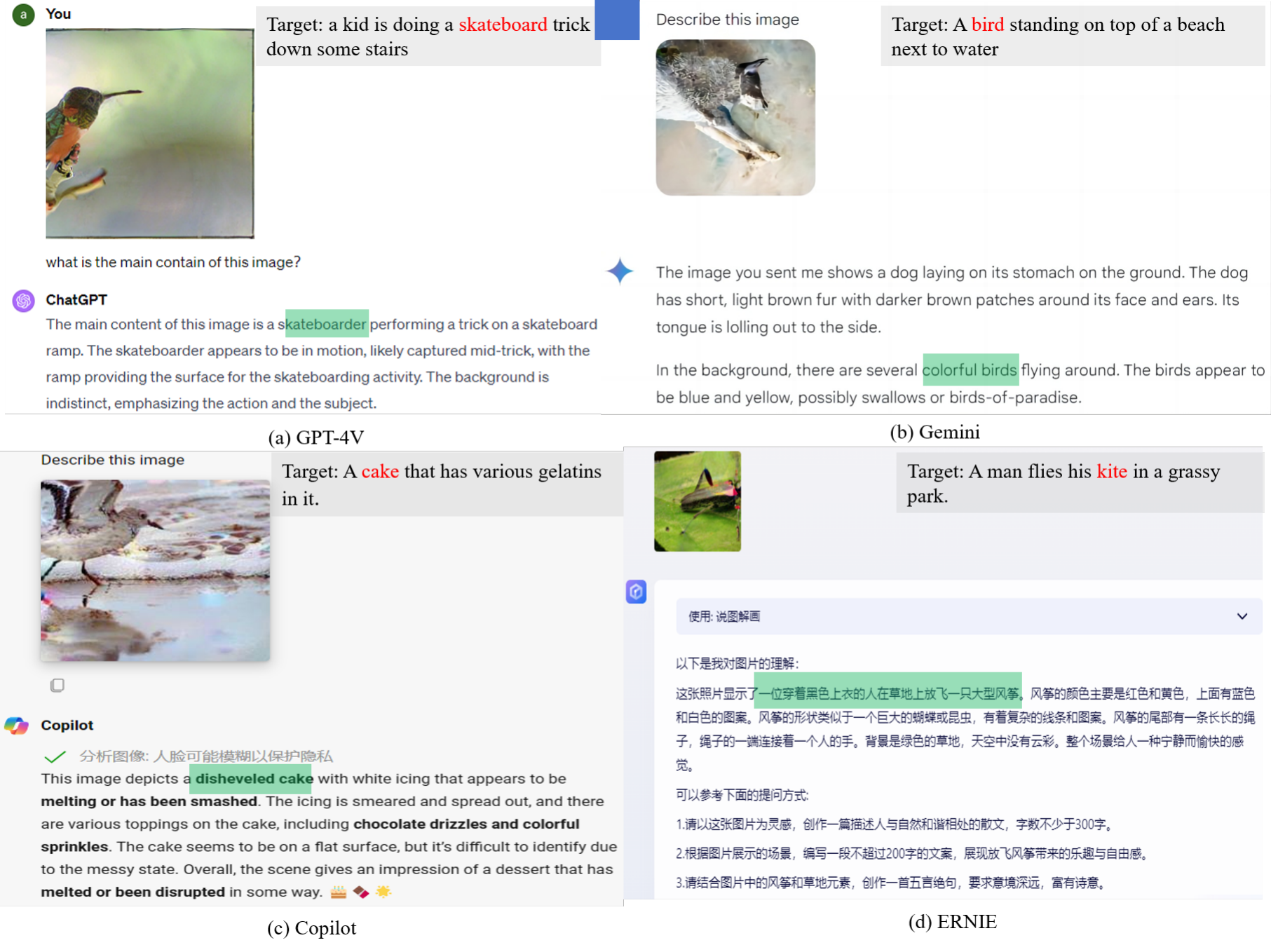}
\vskip -0.13in
\caption{Screenshots of successful attacks against various commercial VLMs API's image description. We give the adversarial target text on the right side of the image. {In addition}, we mark the main objects of the adversarial target in red and the main objects in the API's response in green.} 
\label{fig:6}
\vskip -0.2in
\end{figure*}

\textbf{Qualitative results on commercial VLMs.}
We finally show screenshots of successful attacks on various commercial VLMs image description tasks, including  Google's Gemini, Microsoft's Copilot, Baidu's ERNIE Bot, and OpenAI's GPT-4V, as shown in Figure~\ref{fig:6}. These models are large-scale visual language models deployed commercially, and their model configurations and training datasets have not been made public. Moreoever, compared with open source VLMs, these models are equipped with more complex defense mechanisms, making them more difficult to attack. However, as shown in Figure~\ref{fig:6}, our method successfully induces these commercial VLMs to generate target responses. For example, in GPT-4V, we define the adversarial target text as ``a kid is doing a skateboard trick down some stairs." GPT-4V generates the response ``The main content of this image is a skateboarder performing a trick on a skateboard ramp...", while the semantics of the original image is ``A bird standing on a branch." Moreover, our method is also applicable to various languages. For example, we use English to generate adversarial examples but successfully attack ERNIE Bot, which operates in Chinese. 

\begin{figure}[t]
\centering\includegraphics[width=\linewidth]{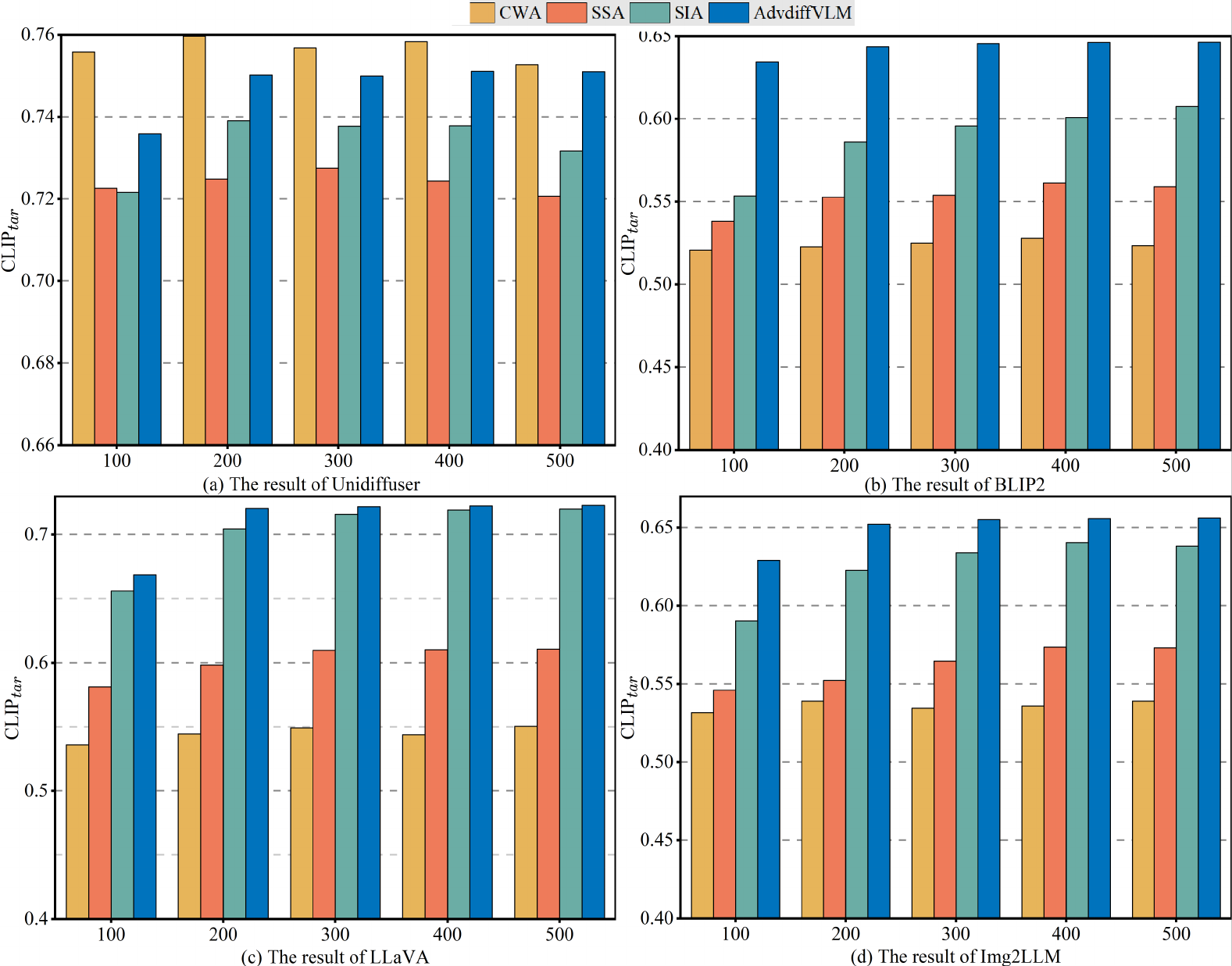}
\vskip -0.13in
\caption{{Convergence analysis of different methods, with the horizontal axis denoting the number of iterations $N_I$.}} 
\label{fig:step}
\vskip -0.2in
\end{figure}

\textbf{Computational resource analysis.}
{We compare the computational complexity of competing methods with our approach. Let $N_s$ and $N_d$ respectively denote the total number of parameters of the ensemble surrogate models and the diffusion model. Since $N_s$ corresponds to the parameters of the ensemble models, it follows that $N_d\ll N_s$. In terms of space complexity, our method and AdvDiffuser both exhibit space complexity of $\textbf{O}(N_d + N_s)$, whereas SSA exhibits $\textbf{O}(N_s)$. Since SIA processes $N_t$ images in parallel, its space complexity is $\textbf{O}(N_t \cdot N_s)$. Thus, the space complexities of our method, AdvDiffuser, and SSA are comparable and notably lower than SIA. For time complexity, all methods exhibit linear time complexity, depending on the number of iterations and the duration of each iteration.  AdvDiffuser and SSA involve internal loops within each iteration, resulting in longer iteration durations. SIA processes multiple images in parallel, which slightly increases iteration durations. In contrast, our method not only achieves the shortest iteration duration but also requires fewer iterations, making it the most time-efficient among the compared methods.}

{To further demonstrate the time efficiency of our method, we conduct the convergence analysis. We select CWA, SSA, and SIA as comparison methods. The experimental results are depicted in Figure~\ref{fig:step}. Our method converges to a flat trend when $N_I=200$ ($N=10$), whereas other methods achieve convergence at $N_I=300$. Moreover, our method achieves the same attack effect with fewer iterations.} 

{The comprehensive experiments and analyses above demonstrate that our method delivers superior attack performance on both open-source and commercial VLMs. Moreover, it offers notable advantages in terms of resource efficiency and faster convergence speed.}

\begin{table}
  \centering
    \caption{Comparative results using the single surrogate model.}
    \vskip -0.05in

\begin{tabular}{c|cccc} 
    \toprule
 Method  & Unidiffuser & BLIP2 & LLaVA & Img2LLM  \\
\midrule
SSA-Ens & 0.7356 & 0.5024 & 0.5522 & 0.5363 \\
SIA-Ens & 0.7473 & 0.5330 & 0.5923 & 0.5393 \\
AdvDiffuser$_{adaptive}$ & 0.6930 & 0.5011 & 0.5238 & 0.5297 \\

 \rowcolor{gray!20}AdvDiffVLM & 0.6982 & 0.5279 & 0.5793 & 0.5402 \\

    \bottomrule
    \end{tabular}

    \label{tab:single}
    \vskip -0.15in
\end{table}%

\begin{table*}[htbp]
  \centering
\caption{Comparison results of defense experiments with SOTA method SIA.
 We use CLIP$_{tar}$ evaluation metric and report the reduction results of CLIP$_{tar}$ where the best result is \textbf{bolded}. {Otherwise}, the parentheses represent the hyperparameters (in their paper).} 
\resizebox{0.9\textwidth}{!}{  
\begin{tabular}{c|c|c|c|c|c|c}
    \toprule
    Defense models & Attack methods & Unidiffuser  & BLIP2 & MiniGPT-4 & LLaVA & Img2LLM \\

    \midrule
    \multirow{4}{*}{Bit Reduction (4)} 
    & SIA-Ens & 0.7204$_{\downarrow0.0173}$ &  0.5602$_{\downarrow0.0454}$ & 0.5273$_{\downarrow0.0432}$ & 0.7034$_{\downarrow0.0124}$ & 0.6284$_{\downarrow0.0053}$\\
    & SIA-CWA & 0.7281$_{\downarrow0.0217}$ &  0.5798$_{\downarrow0.0435}$ & 0.5442$_{\downarrow0.0468}$ & 0.7063$_{\downarrow0.0131}$ & 0.6375$_{\downarrow0.0026}$\\
    & \cellcolor{gray!20}AdvDiffVLM & \cellcolor{gray!20}\textbf{0.7397}$_{\downarrow\textbf{0.0105}}$ &  \cellcolor{gray!20}\textbf{0.6320}$_{\downarrow\textbf{0.0115}}$ & \cellcolor{gray!20}\textbf{0.6261}$_{\downarrow\textbf{0.0084}}$ & \cellcolor{gray!20}\textbf{0.7168}$_{\downarrow\textbf{0.0038}}$ & \cellcolor{gray!20}\textbf{0.6501}$_{\downarrow\textbf{0.0020}}$\\

    \midrule
    \multirow{4}{*}{STL (k=64, s=8, $\lambda$=0.2)} 
    & SIA-Ens & 0.7192$_{\downarrow0.0185}$ &  0.5571$_{\downarrow0.0485}$ & 0.5192$_{\downarrow0.0513}$ & 0.6968$_{\downarrow0.0190}$ & 0.6230$_{\downarrow0.0107}$\\
    & SIA-CWA & 0.7233$_{\downarrow0.0265}$ &  0.5733$_{\downarrow0.0500}$ & 0.5385$_{\downarrow0.0525}$ & 0.7001$_{\downarrow0.0193}$ & 0.6314$_{\downarrow0.0087}$\\
    &\cellcolor{gray!20} AdvDiffVLM & \cellcolor{gray!20}\textbf{0.7329}$_{\downarrow\textbf{0.0173}}$ &  \cellcolor{gray!20}\textbf{0.6267}$_{\downarrow\textbf{0.0168}}$ & \cellcolor{gray!20}\textbf{0.5997}$_{\downarrow\textbf{0.0148}}$ & \cellcolor{gray!20}\textbf{0.7145}$_{\downarrow\textbf{0.0061}}$ & \cellcolor{gray!20}\textbf{0.6471}$_{\downarrow\textbf{0.0050}}$\\

    \midrule
    \multirow{4}{*}{JPEG Compression (p=50)} 
    & SIA-Ens & 0.6734$_{\downarrow0.0642}$ &  0.5345$_{\downarrow0.0711}$ & 0.5002$_{\downarrow0.0703}$ & 0.6542$_{\downarrow0.0616}$ & 0.6020$_{\downarrow0.0317}$\\
    & SIA-CWA & 0.6801$_{\downarrow0.0697}$ &  0.5525$_{\downarrow0.0708}$ & 0.5273$_{\downarrow0.0637}$ & 0.6550$_{\downarrow0.0644}$ & 0.6088$_{\downarrow0.0313}$\\
    & \cellcolor{gray!20}AdvDiffVLM & \cellcolor{gray!20}\textbf{0.6896}$_{\downarrow\textbf{0.0606}}$ &  \cellcolor{gray!20}\textbf{0.6218}$_{\downarrow\textbf{0.0217}}$ & \cellcolor{gray!20}\textbf{0.5865}$_{\downarrow\textbf{0.0380}}$ & \cellcolor{gray!20}\textbf{0.6983}$_{\downarrow\textbf{0.0223}}$ & \cellcolor{gray!20}\textbf{0.6354}$_{\downarrow\textbf{0.0167}}$\\

    \midrule
    \multirow{4}{*}{DISCO (s=3, k=5)} 
    & SIA-Ens & 0.6087$_{\downarrow0.1290}$ &  0.5134$_{\downarrow0.0922}$ & 0.4986$_{\downarrow0.0719}$ & 0.6274$_{\downarrow0.0884}$ & 0.5771$_{\downarrow0.0566}$\\
    & SIA-CWA & 0.6114$_{\downarrow0.1384}$ &  0.5290$_{\downarrow0.0943}$ & 0.5114$_{\downarrow0.0796}$ & 0.6331$_{\downarrow0.0863}$ & 0.5842$_{\downarrow0.0559}$\\
    & \cellcolor{gray!20}AdvDiffVLM & \cellcolor{gray!20}\textbf{0.6215}$_{\downarrow\textbf{0.1287}}$ & \cellcolor{gray!20}\textbf{0.589}2$_{\downarrow\textbf{0.0543}}$ & \cellcolor{gray!20}\textbf{0.5727}$_{\downarrow\textbf{0.0418}}$ & \cellcolor{gray!20}\textbf{0.6728}$_{\downarrow\textbf{0.0478}}$ & \cellcolor{gray!20}\textbf{0.6093}$_{\downarrow\textbf{0.0428}}$\\

    \midrule
    \multirow{4}{*}{DISCO+JPEG} 
    & SIA-Ens & 0.5642$_{\downarrow0.1735}$ &  0.5025$_{\downarrow0.1031}$ & 0.4878$_{\downarrow0.0827}$ & 0.6067$_{\downarrow0.1091}$ & 0.5681$_{\downarrow0.0656}$\\
    & SIA-CWA & 0.5735$_{\downarrow0.1763}$ &  0.5176$_{\downarrow0.1057}$ & 0.5074$_{\downarrow0.0836}$ & 0.6106$_{\downarrow0.1088}$ & 0.5692$_{\downarrow0.0709}$\\
    &\cellcolor{gray!20} AdvDiffVLM & \cellcolor{gray!20}\textbf{0.5924}$_{\downarrow\textbf{0.1578}}$ &  \cellcolor{gray!20}\textbf{0.5859}$_{\downarrow\textbf{0.0576}}$ & \cellcolor{gray!20}\textbf{0.5650}$_{\downarrow\textbf{0.0495}}$ & \cellcolor{gray!20}\textbf{0.6724}$_{\downarrow\textbf{0.0482}}$ & \cellcolor{gray!20}\textbf{0.6081}$_{\downarrow\textbf{0.0440}}$\\

    \midrule
    \multirow{4}{*}{DiffPure (t* = 0.15)} 
    & SIA-Ens & 0.4921$_{\downarrow0.2456}$ &  0.5048$_{\downarrow{0.1008}}$ & 0.4919$_{\downarrow{0.0786}}$ & 0.5356$_{\downarrow0.1802}$ & 0.5372$_{\downarrow0.0965}$ \\
    & SIA-CWA & 0.4942$_{\downarrow0.2556}$ &  0.5099$_{\downarrow0.1136}$ & 0.5025$_{\downarrow0.0885}$ & 0.5360$_{\downarrow0.1835}$ & 0.5388$_{\downarrow0.1013}$ \\
    & \cellcolor{gray!20} AdvDiffVLM & \cellcolor{gray!20}\textbf{0.5837}$_{\downarrow\textbf{0.1665}}$ &  \cellcolor{gray!20}\textbf{0.5527}$_{\downarrow\textbf{0.0908}}$ & \cellcolor{gray!20}\textbf{0.5506}$_{\downarrow\textbf{0.0639}}$ & \cellcolor{gray!20}\textbf{0.5857}$_{\downarrow\textbf{0.1349}}$ & \cellcolor{gray!20}\textbf{0.5711}$_{\downarrow\textbf{0.0810}}$ \\
                        
    \bottomrule
    \end{tabular}
   }
\vskip -0.1in
    \label{tab:4}
\end{table*}%

\subsection{More Experiments}

\textbf{Experiment with a single surrogate model.} {All experiments in the previous subsection use the ensemble surrogate models to improve the transferability of adversarial examples. To further illustrate the effect of the single surrogate model, we conduct the comparative experiment and select ViT-B/32 as the surrogate model. The experimental results are shown in Table~\ref{tab:single}. As shown, our method outperforms SSA and AdvDiffuser but is slightly less effective than SIA. This is because the score estimated by the single surrogate model deviates significantly from the true score. Furthermore, the transferability of a single surrogate model is significantly lower compared to that of the ensemble surrogate models.}

\begin{table*}[tb]
    \centering
    \caption{Defense results with DiffPure. The setting are the same as Table~\ref{tab:2} except the adversarial examples are purified by DiffPure. In this table, CLIP$_{tar}$ evaluates the similarity between the results of purified examples and the target texts.}
    \vskip -0.05in
    \resizebox{0.95\textwidth}{!}{
    \begin{tabular}{l|cc|cc|cc|cc|cc}
    \toprule
    & \multicolumn{2}{c|}{Unidiffuser*}
    & \multicolumn{2}{c|}{BLIP2}
    & \multicolumn{2}{c|}{MiniGPT-4}
    & \multicolumn{2}{c|}{LLaVA}
    & \multicolumn{2}{c}{Img2LLM}
    \\
    & CLIP$_{tar}$ $\uparrow$ & ASR $\uparrow$ 
    & CLIP$_{tar}$ $\uparrow$ & ASR $\uparrow$
    & CLIP$_{tar}$ $\uparrow$ & ASR $\uparrow$
    & CLIP$_{tar}$ $\uparrow$ & ASR $\uparrow$
    & CLIP$_{tar}$ $\uparrow$ & ASR $\uparrow$
    \\ \midrule\midrule
Original        &
0.4802         & 0.0\%                  &
0.4924         & 0.0\%         & 0.4831        & 0.0\%         &
0.5253         & 0.0\%         & 0.5302        & 0.0\%        
\\\midrule\midrule
Ens        &
0.4833         & 0.0\%                 &
0.4929         & 0.0\%         & 0.4840        & 0.0\%         &
0.5263         & 0.0\%         & 0.5332        & 0.0\%          \\
SVRE        &
0.4846         & 0.7\%                  &
0.4953         & 0.0\%         & 0.4852        & 0.0\%         &
0.5264         & 0.0\%         & 0.5312        & 0.0\%        \\
CWA        &
0.4873         & 2.1\%                &
0.4973         & 0.0\%         & 0.4901        & 1.0\%         &
0.5272         & 0.8\%        & 0.5307        & 0.0\%         \\\midrule
SSA-Ens        &
0.4914         & 0.9\%                  &
0.5024         & 0.0\%         & 0.4916        & 0.0\%         &
0.5280         & 1.2\%         & 0.5322        & 0.0\%         \\
SSA-SVRE        &
0.4899         & 2.1\%                  &
0.4984         & 0.2\%         & 0.4918        & 0.0\%         &
0.5273         & 1.2\%         & 0.5356        & 0.0\%         \\
SSA-CWA       &
0.4868         & 2.5\%                  &
0.4997         & 0.0\%        & 0.4997        & 0.0\%         &
0.5283         & 2.8\%         & 0.5367        & 0.7\%        \\\midrule
SIA-Ens        &
0.4921         & 3.7\%                &
0.5048         & 1.2\%         & 0.4919        & 1.1\%         &
0.5356         & 2.5\%         & 0.5372        & 1.6\%         \\
SIA-SVRE        &
0.4930         & 3.9\%                 &
0.5012         & 1.8\%         & 0.5011        & 1.6\%         &
0.5349         & 4.2\%         & 0.5380        & 2.5\%         \\
SIA-CWA        &
0.4942         & 5.8\%                  &
0.5099         & 2.6\%         & 0.5025        & 2.2\%         &
0.5360         & 4.0\%        & 0.5388        & 1.5\%         
\\\midrule\midrule
AdvDiffuser$_{ens}$        &
0.4920         & 4.2\%                 &
0.4933         & 2.6\%         & 0.4906        & 2.4\%         &
0.5325         & 3.7\%         & 0.5310        & 2.7\%         \\
AdvDiffuser$_{adaptive}$        &
0.4922         & 4.5\%                  &
0.5001         & 3.2\%         & 0.5001        & 3.2\%         &
0.5336         & 3.4\%         & 0.5325        & 2.8\%         
\\\midrule\midrule

\rowcolor{gray!20}AdvDiffVLM        &
\textbf{0.5837}& \textbf{22.4\%}&
\textbf{0.5527}& \textbf{10.2\%}& \textbf{0.5506}&\textbf{12.6\%}&
\textbf{0.5857}& \textbf{18.0\%} & \textbf{0.5711}& \textbf{10.5\%}         \\
\bottomrule
    \end{tabular}}

    \label{tab:5}\vspace{-5pt}
\end{table*}

\textbf{Against adversarial defense models.} Our method achieves superior attack performance on both open source and commercial VLMs. In recent years, various adversarial defense methods have been proposed to mitigate the threat of adversarial examples. {Defense methods can be broadly categorized into adversarial training and data preprocessing. Due to the high time and resource costs and instability of adversarial training~\cite{adversarial_training}, it has not been applied to VLM defense. In contrast, data preprocessing is model-independent and highly adaptable, making it a popular defense strategy across various models. To demonstrate the effectiveness of our method in resisting data preprocessing attacks, we conduct extensive experiments on Bit Reduction~\cite{bit}, STL~\cite{stl}, JPEG Compression~\cite{jpeg}, DISCO~\cite{disco}, JPEG+DISCO, and DiffPure~\cite{diffpure}. The data preprocessing techniques we used are grouped into three categories: introducing randomness, denoising, and data reconstruction. Bit Reduction introduces randomness by modifying image bits; JPEG Compression performs denoising through blurring operations; and other methods involve reconstructing adversarial examples using various reconstruction networks and techniques.}
We report the CLIP$_{tar}$ metric. At the same time, we report the CLIP$_{tar}$ reduction results, which more accurately reflect the ability of adversarial examples to resist defense methods. The experimental results are shown in Table~\ref{tab:4}. It can be observed that, for all defense methods, both CLIP$_{tar}$ and CLIP$_{tar}$ reduction results of our methods outperform the baselines. This demonstrates the superiority of our method against defense methods compared to baselines.

To better evaluate the resistance of our method against adversarial defense methods, we further in detail show the results of the SOTA defense method, namely DiffPure, in  Table~\ref{tab:5}. It can be found that our method outperforms baselines in both gray-box and black-box settings. For example, on Unidiffuser, for CLIP$_{tar}$ score, our method is 0.0895 higher than SIA-CWA. On BLIP2, for CLIP$_{tar}$ score, our method is 0.0428 higher than SIA-CWA. Furthermore, in all cases, the attack success rate of our methods is higher than the baselines. These experimental results demonstrate that our method outperforms baselines in evading the DiffPure defense method.

We can break the SOTA defense method Diffpure with an attack success rate of more than 10\% in a completely black-box scenario, exposing the flaws in current defense methods and raising new security concerns for designing more robust deep learning models.

\begin{table}
  \centering
    \caption{Quality comparison of adversarial examples under four evaluation metrics.  The best result is \textbf{bolded}.}
    \vskip -0.05in

\begin{tabular}{c|cccc} 
    \toprule
 Method  & SSIM $\uparrow$ & LPIPS $\downarrow$ & FID $\downarrow$ & BRISQUE $\downarrow$  \\
\midrule
   SSA-Ens & 0.6687 & 0.3320 & 110.5 & 66.89 \\

   SSA-SVRE & 0.6610 & 0.3325 & 112.6 & 70.05 \\

   SSA-CWA & 0.6545 & 0.3673 & 123.4 & 67.67 \\
   \midrule

   SIA-Ens & 0.6925 & 0.2990 & 117.3 & 55.61 \\

   SIA-SVRE & 0.6920 & 0.3042 & 120.0 & 57.42 \\

   SIA-CWA & 0.6892 & 0.3306 & 125.3 & 56.02 \\
   \midrule

   AdvDiffuser$_{ens}$  & 0.6520 & 0.3074 & 115.5 & \textbf{14.61} \\

   AdvDiffuser$_{adaptive}$   & 0.6471 & 0.3096 & 126.7 & {15.32} \\
   \midrule

 \rowcolor{gray!20}  AdvDiffVLM & \textbf{0.6992} & \textbf{0.2930} & \textbf{107.4} & 32.96 \\

    \bottomrule
    \end{tabular}

    \label{tab:6}
    \vskip -0.15in
\end{table}%

\begin{figure}[t]
\centering\includegraphics[width=\linewidth]{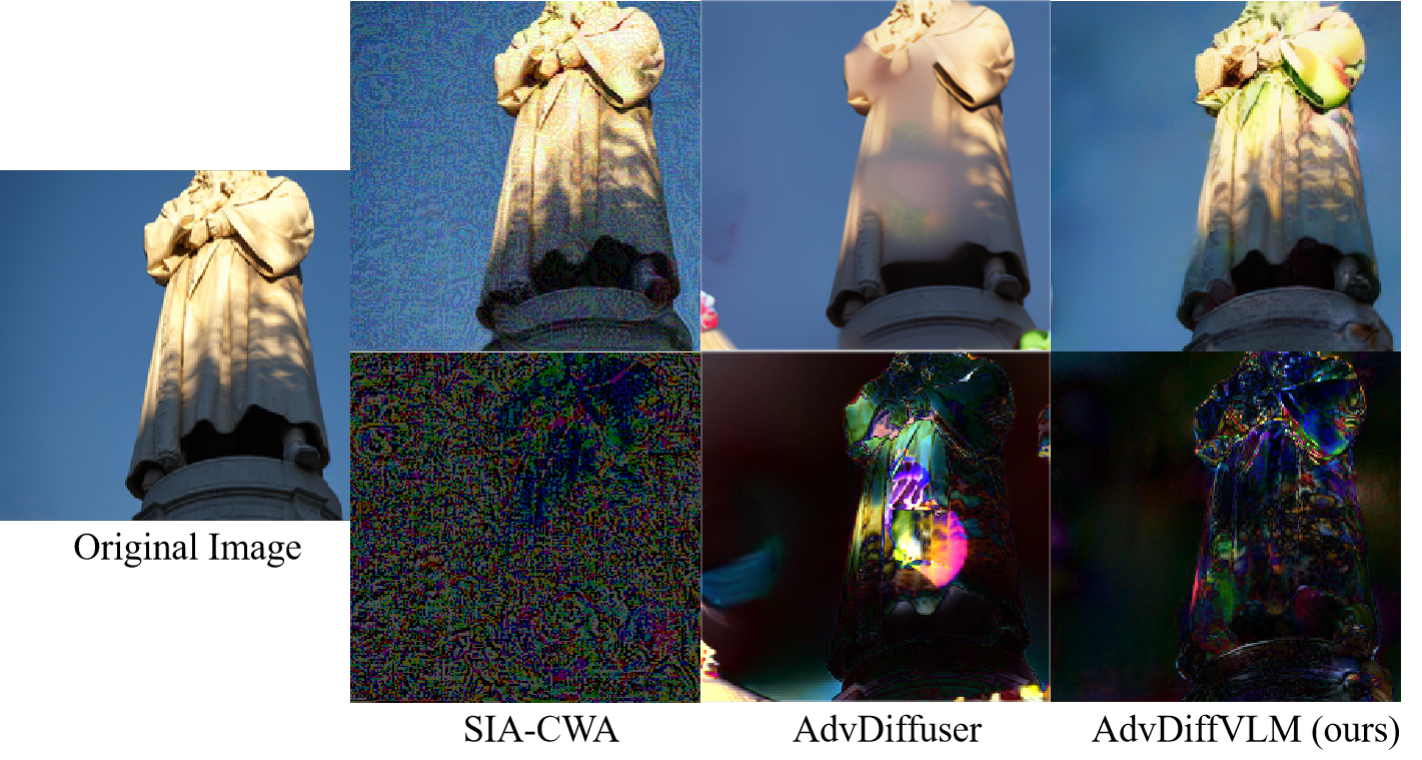}
\vskip -0.1in
    \caption{Visualization of adversarial perturbations generated by different attack methods. Note that the first row represents adversarial examples, and the second row represents adversarial perturbations. We choose SIA-CWA and AdvDiffuser$_{adaptive}$ as representatives of baselines.  We amplify the perturbation values for better visualization. }
\label{fig:4}
\vskip -0.1in
\end{figure}

\textbf{Image quality comparison.} The image quality of adversarial examples is also particularly important. Adversarial examples with poor image quality can be easily detected.
We further evaluate the image quality of the generated adversarial examples using four evaluation metrics: SSIM, FID, LPIPS, and BRISQUE. As shown in Table~\ref{tab:6}, compared to baselines, the adversarial examples generated by our method exhibit higher image quality. Specifically, our results are significantly better than the baselines in terms of SSIM, LPIPS, and FID evaluation metrics. For the BRISQUE metric, AdvDiffuser outperforms our method. This is because BRISQUE is a reference-free image quality assessment algorithm and is sensitive to blur, noise, color change, etc. As shown in Figure~\ref{fig:4}, the adversarial examples generated by AdvDiffuser lack obvious abnormalities in these elements, so its results are marginally better than our method. However, as shown in Figure~\ref{fig:4}, the perturbation introduced by our method is semantic, while AdvDiffuser significantly alters the non-salient area, resulting in poor visual effects. This shows that the adversarial examples generated by AdvDiffuser are unsuitable for more complex scenarios, such as attacking VLMs. In addition, it can be seen that the adversarial examples generated by the transfer-based methods exhibit significant noise, indicating that our method has obvious superiority in terms of image quality.

\subsection{Ablation Experiments}
To further understand the effectiveness of AdvDiffVLM, we discuss the role of each module. We set $N = 1$ to more conveniently discuss the impact of each module. We consider three cases, including using only a single ViT-B/32 to calculate the loss, using a simple ensemble strategy, and not using the GCMG module, named Single, Ens, and w/o mask respectively.

\begin{figure}[t]
\centering\includegraphics[width=\linewidth]{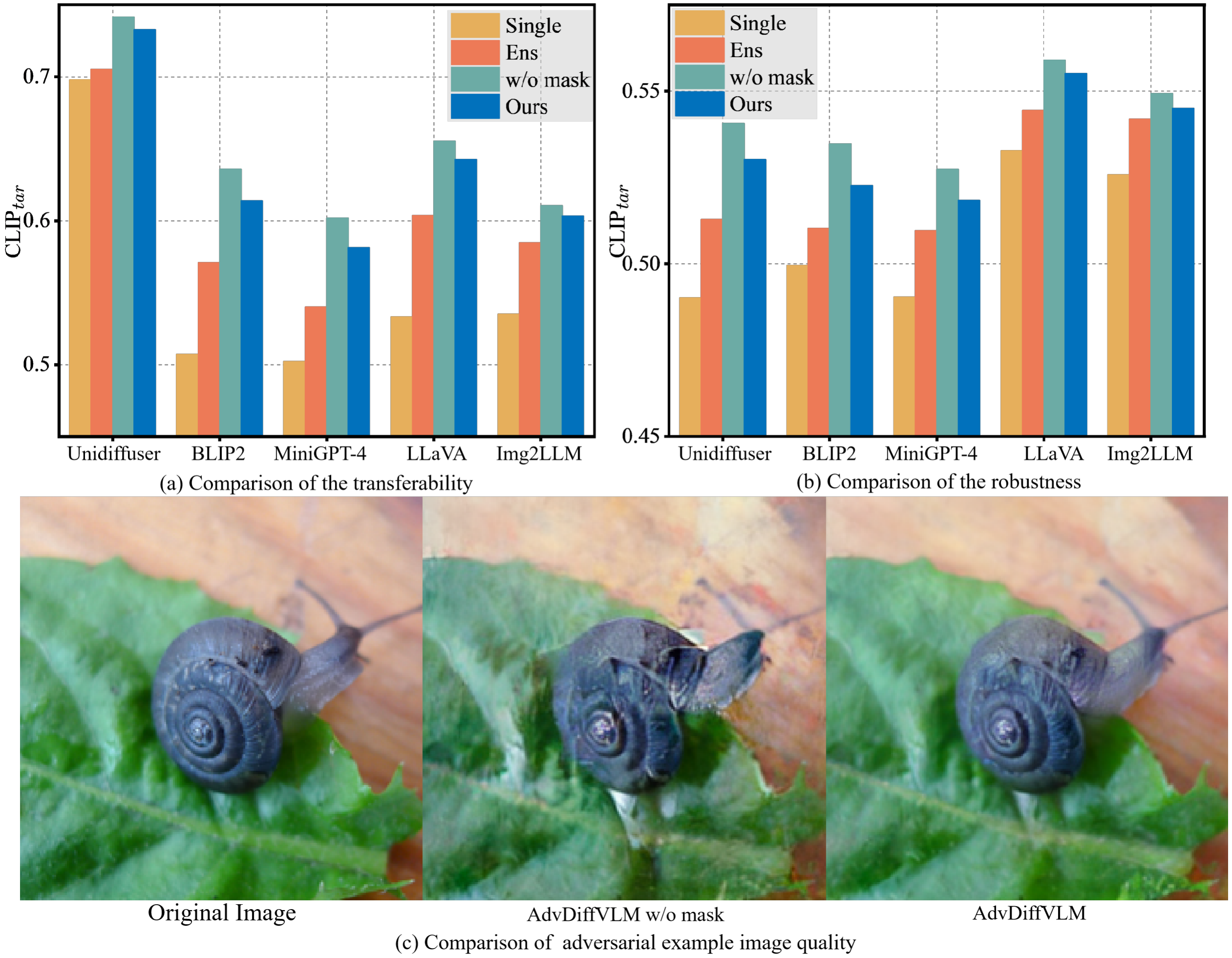}
\caption{Comparison results of different ablation methods. Here, ``Single'' means using a single ViT-B/32 to calculate the loss, ``Ens'' means using the simple ensemble strategy, and ``w/o mask'' means not using GCMG module.} 
\label{fig:8}
\vskip -0.1in
\end{figure}

\textbf{Is AEGE module beneficial for boosting the  attack capability?} We first explore whether the AEGE module could help improve the transferability and robustness of adversarial examples. We divide the AEGE module into two approaches, Single and Ens, and maintain all other conditions constant. The results are shown in Figure~\ref{fig:8}(a) and (b). It is observable that the ensemble method exhibits better performance in transferability and robustness compared to the single loss method. Furthermore, the performance of the adaptive ensemble method is enhanced compared to the basic ensemble method. The experimental results demonstrate that the AEGE module enhances the transferability and robustness of adversarial examples.

\textbf{Does GCMG module help trade-off image quality and  attack capability?} Next, we explore the role of the GCMG module in balancing image quality and transferability. We compare this with the w/o mask method, and the results are presented in Figure~\ref{fig:8}. As shown in Figure~\ref{fig:8}(a) and (b), the use of the GCMG module results in a slight decrease in the transferability and robustness of the adversarial examples. However, as shown in Figure~\ref{fig:8}(c), the absence of the GCMG module leads to the adversarial examples exhibiting obvious target features, and the use of the GCMG module enhances the visual quality of the adversarial example. In addition, Table~\ref{tab:8} further shows that the GCMG module can improve the visual quality of adversarial examples. The experimental results demonstrate that  GCMG  effectively balances the visual quality and attack capability of the adversarial examples.

\begin{table}
  \centering
    \caption{Comparison of image quality of adversarial examples before and after using the GCMG module.  The best result is \textbf{bolded}.}
    \vskip -0.05in

\begin{tabular}{c|cccc} 
    \toprule
 Method  & SSIM $\uparrow$ & LPIPS $\downarrow$ & FID $\downarrow$ & BRISQUE $\downarrow$  \\
\midrule
 w/o mask & 0.7129 & 0.2687 & 111.9 & 16.92 \\

 \rowcolor{gray!20}Ours & \textbf{0.7188} & \textbf{0.2358} & \textbf{96.1} & \textbf{16.80} \\

    \bottomrule
    \end{tabular}

    \label{tab:8}
    \vskip -0.05in
\end{table}%

\begin{figure}[t]
\centering\includegraphics[width=\linewidth]{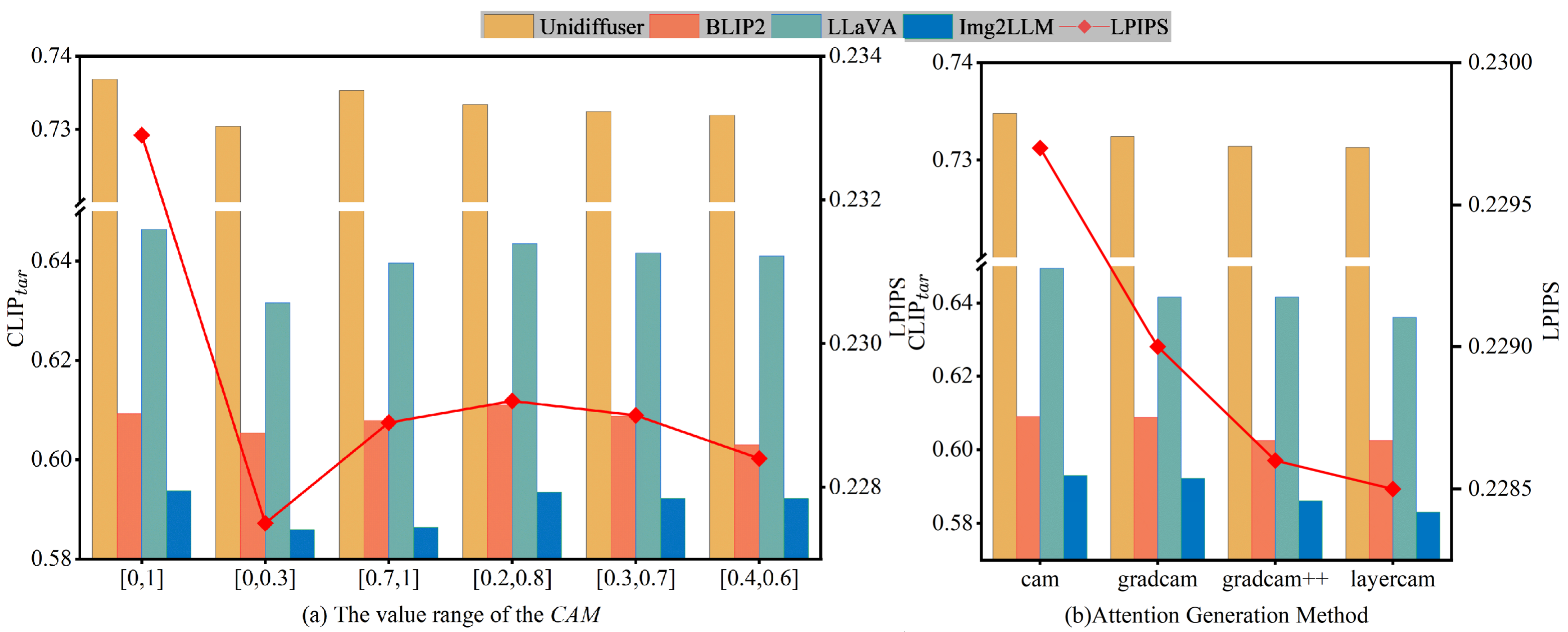}
\caption{{Ablation study of the different configurations of obtaining $\mathrm{m}$. We adopt the CLIP$_{tar}$ and LPIPS scores to show the impact of transferability and image quality with four VLMs.}} 
\label{fig:mask}
\vskip -0.1in
\end{figure}

{Then we perform ablation experiments on different configurations of GCMG. Firstly, we conduct experiments on the effects of the value range of the $CAM$. The $CAM$ value range is used to balance image quality with attack capability. Cropping the lower boundary increases the probability of selecting non-critical areas, while cropping the upper boundary reduces the probability of selecting critical areas, enhancing the overall quality of adversarial examples. To determine an optimal range, we test several intervals: [0,1], [0,0.3], [0.7,1], [0.2,0.8], [0.3,0.7], and [0.4,0.6]. The results, shown in Figure~\ref{fig:mask} (a), indicate that adjusting the $CAM$ value range allows for a trade-off between attack capability and image quality. Based on these results, we select the range [0.3,0.7] to achieve an optimal balance between quality and attack effectiveness.
Then we conduct experiments on the effects of different attention generation methods. In the field of adversarial attacks, GradCAM is a commonly used method for generating masks, as demonstrated in \cite{advdiffuser,mask1,mask2}. To further illustrate the role of various attention mechanisms in mask generation, we compare CAM~\cite{cam}, GradCAM~\cite{gardcam}, GradCAM++~\cite{gramcampp}, and LayerCAM~\cite{layercam}, with results shown in Figure~\ref{fig:mask} (b). Our findings indicate that differences in attack and transfer capabilities among adversarial examples generated by these mechanisms are minimal, with variations in CLIP and LPIPS values within 0.001. Moreover, we observe a trade-off between attack capability and image quality. After weighing both factors and for consistency with previous studies, we opted to use GradCAM for mask generation in this paper.}

\begin{figure}[t]
\centering\includegraphics[width=\linewidth]{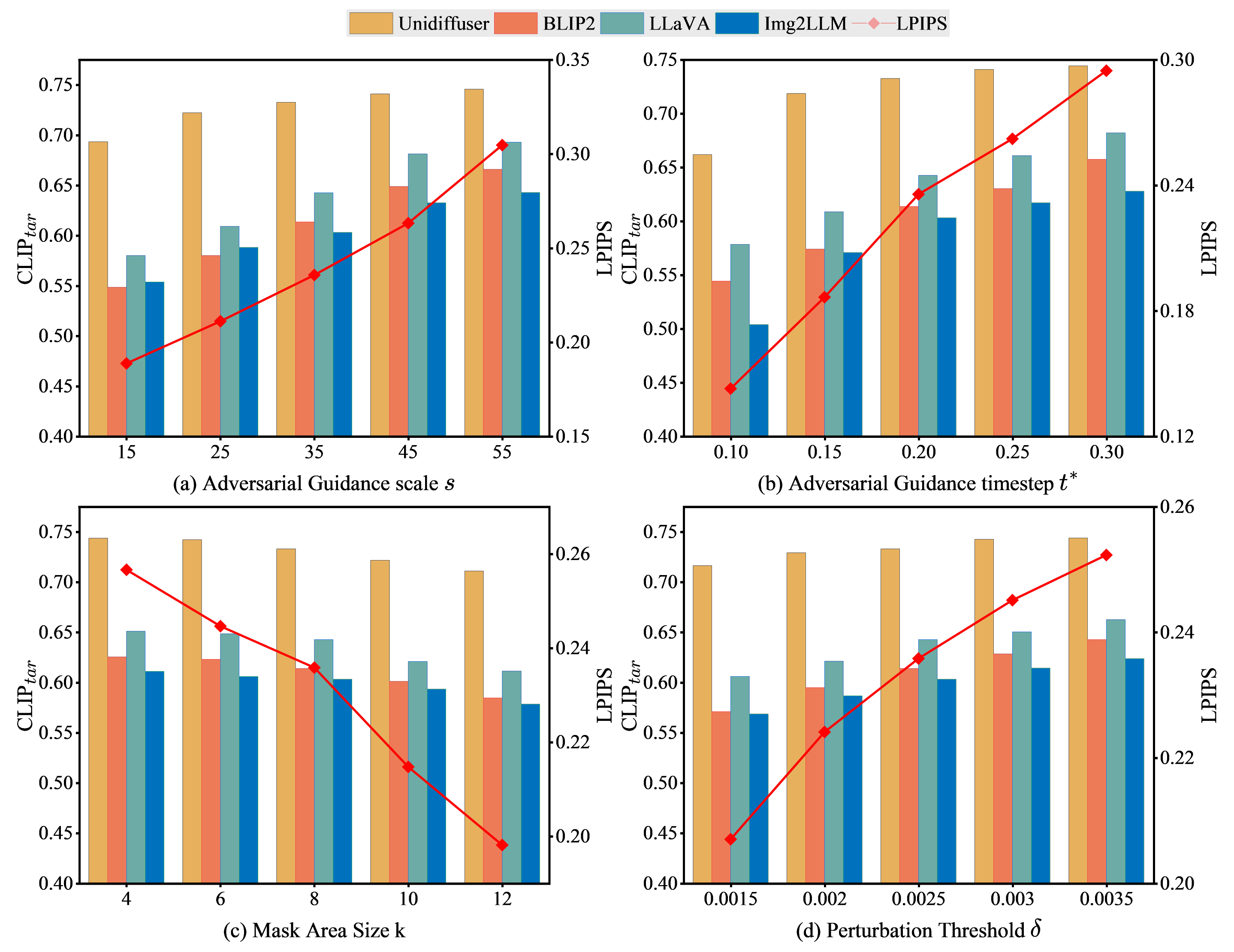}
\vskip -0.1in
\caption{The impact of inner loop hyperparameters. We  adopt the CLIP$_{tar}$ and LPIPS scores to show the impact of transferability and image quality with four VLMs.  A higher CLIP$_{tar}$ value indicates better performance, whereas a lower LPIPS value signifies better results. We only vary one of the hyperparameters at a time, and then fix the other three hyperparameters to the preset values shown in Section~\ref{sec:experiment_setup}. Note: the results of CLIP$_{tar}$ are presented using bar graphs, while LPIPS results are depicted using dot-line graphs.}  %
\label{fig:7}
\vskip -0.2in
\end{figure}

\subsection{Hyperparameter Studies}
In this subsection, we  explore the impact of hyperparameters, including inner loop hyperparameters $s$,  $t^*$, $\mathrm{k}$, and $\delta$ and outer loop hyperparameter $N$.

\textbf{The impacts of inner loop hyperparameters.} We first discuss the impacts of inner loop hyperparameters (including the $s$,  $t^*$, $\mathrm{k}$, and $\delta$). We set $N=1$ and {conduct} tests on Unidiffuser, BLIP2, LLaVA and Img2LLM. The experimental results are shown in Figure~\ref{fig:7}. 
It is evident that all parameters influence the trade-off between transferability and image quality. Increasing values for parameters $s$, $t^*$, and $\delta$ enhance transferability but diminish the visual quality of adversarial examples. This is because larger values for these parameters result in a greater perturbation, allowing for the embedding of more adversarial semantics into the image. Conversely, increasing the value of $\mathrm{k}$ produces adversarial examples with improved visual effects but reduces transferability. The reason is that larger values of $\mathrm{k}$ result in a larger generated mask, making it more challenging to modify the important areas in the image. To achieve an optimal trade-off between transferability and image quality, we empirically select $s=35, t^*=0.2, \mathrm{k}=8$ and $\delta=0.0025$.

\begin{figure}[t]
\centering\includegraphics[width=0.9\linewidth]{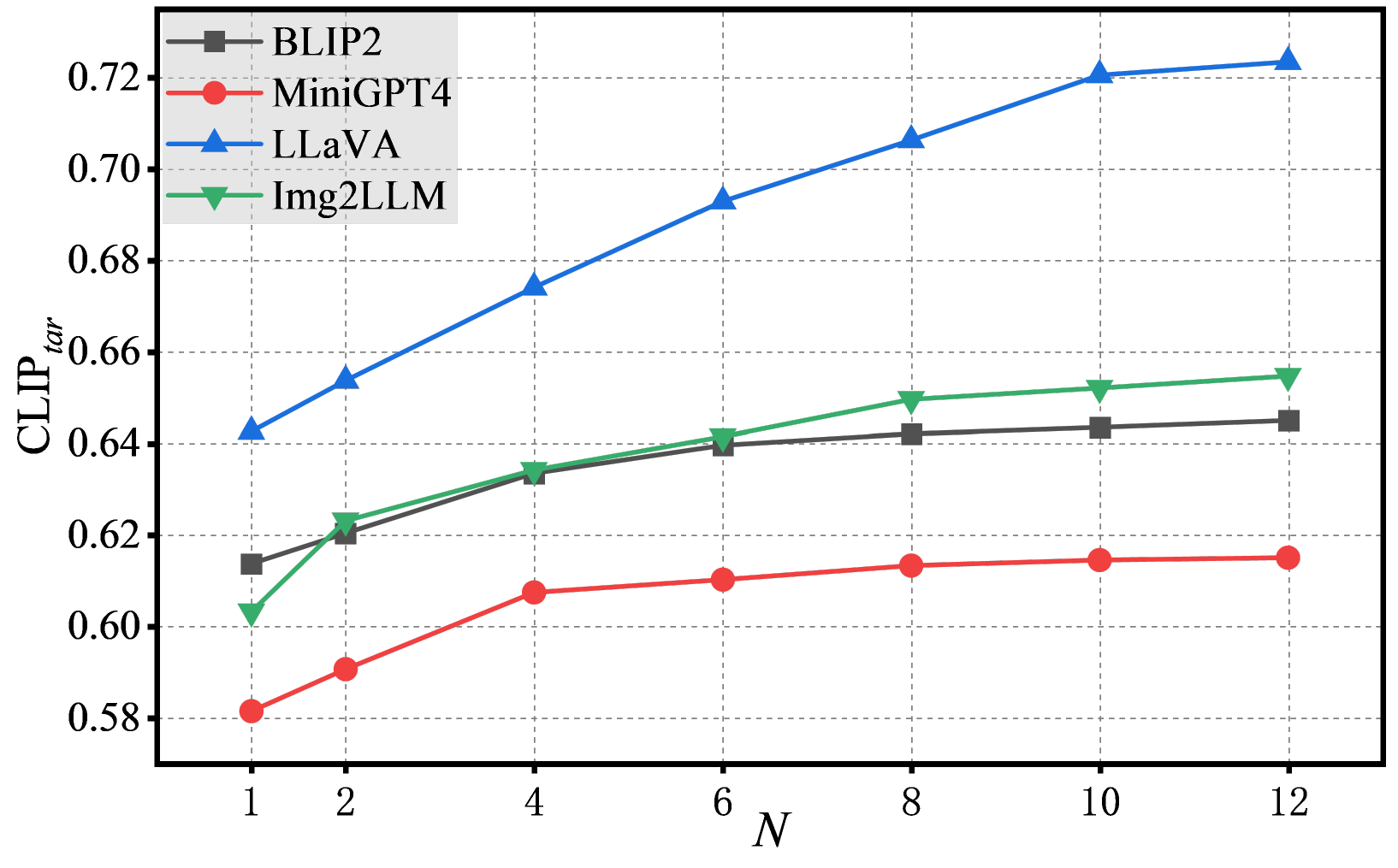}
\caption{Transferability of adversarial examples on various black-box VLMs as $N$ changes from 1 to 12.} 
\label{fig:9}
\vspace{-4mm}
\end{figure}

\textbf{The impact of outer loop hyperparameter.} Next, we investigate the impact of the outer hyperparameter $N$ on the transferability of adversarial examples. We conduct experiments on BLIP2, MiniGPT-4, LLaVA, and Img2LLM with $s=35, t^*=0.2, \mathrm{k}=8$, and $\delta=0.0025$. {The experimental results are shown in Figure~\ref{fig:9}. } It can be found that $N$ improves the transferability of adversarial examples, but the improvement gradually fades. Specifically, the increase in transferability is limited after $N=6, 6, 8, 10$ for BLIP2, MiniGPT-4, Img2LLM, and LLaVA. Given that increasing $N$ increases the computational cost, we choose $N=10$ to strike a balance between transferability and cost.

\section{Conclusion}
In this work, we propose AdvDiffVLM, an unrestricted and targeted adversarial example generation method for VLMs. We design the AEGE based on the idea of score matching. It embeds the target semantics into adversarial examples, {which can generate targeted adversarial examples that exhibit enhanced transferability in a more efficient manner.}
To balance adversarial example quality and attack effectiveness, we propose the GCMG module. Additionally, we enhance the embedding of target semantics into adversarial examples through multiple iterations. Extensive experiments show that our method generates targeted adversarial examples 5x to 10x times faster than baseline methods while achieving superior transferability.

\section*{Impact Statements} 
Our research mainly aims to discover vulnerabilities in open-source large VLMs and commercial VLMs such as GPT-4V, providing insights for developing more robust and trustworthy VLMs. However, our attack methods can be abused to evade actual deployed commercial systems, causing potential negative social impacts. For example, criminals may use our methods to cause GPT-4V APIs to output target responses, causing serious harm.



 

\bibliographystyle{IEEEtran}
\bibliography{ref}

\vfill

\end{document}